\documentclass[10pt,journal,compsoc]{IEEEtran}

\usepackage[switch]{lineno}
\usepackage{epsfig}
\usepackage{graphicx}
\usepackage{amsmath}
\usepackage{amssymb}
\usepackage{wasysym}
\usepackage{tabularx}
\usepackage{xcolor}
\usepackage{threeparttable}
\usepackage{algorithm}
\usepackage{algorithmic}
\usepackage{colortbl}
\usepackage{graphicx}
\usepackage{subfig}
\usepackage{ragged2e}
\usepackage{bm}

\usepackage[pagebackref=false,breaklinks=true,colorlinks,bookmarks=false]{hyperref}

\usepackage{amsfonts}
\usepackage{array}
\usepackage{textcomp}
\usepackage{stfloats}
\usepackage{url}
\usepackage{verbatim}
\usepackage{booktabs, multirow, hhline}
\usepackage{colortbl}
\usepackage{pifont}
\usepackage{threeparttable}
\usepackage{tcolorbox}
\usepackage{arydshln}
\usepackage{tikz}
\usepackage{amssymb}

\newcolumntype{C}[1]{>{\centering\arraybackslash}p{#1}}

\newcommand{\cmark}{\ding{51}}%
\newcommand{\xmark}{\ding{55}}%

\ifCLASSOPTIONcompsoc
  \usepackage[nocompress]{cite}
\else
  \usepackage{cite}
\fi

\ifCLASSINFOpdf
\else
\fi

\usepackage{xspace}
\makeatletter
\newcommand{\thickhline}{%
    \noalign {\ifnum 0=`}\fi \hrule height 0.7pt
    \futurelet \reserved@a \@xhline
}

\newcolumntype{I}{!{\vrule width 0.8pt}}

\DeclareRobustCommand\onedot{\futurelet\@let@token\@onedot}
\def\@onedot{\ifx\@let@token.\else.\null\fi\xspace}

\newcolumntype{P}[1]{>{\centering\arraybackslash}p{#1}}
\definecolor{lightgray}{gray}{.9}
\definecolor{deepgray}{gray}{.8}
\definecolor{mygreen}{RGB}{29, 154, 120}
\definecolor{DarkGreen}{RGB}{42,110,63}

\begin{document}

\title{Generative Retrieval for Unsupervised Text-Based Person Search}

\author{Mang Ye,~\IEEEmembership{Senior Member,~IEEE}, Yucheng Ji, Yang Bai, Min Cao, Siyuan Chai, Bo Du,~\IEEEmembership{Senior Member,~IEEE}, Min Zhang
\IEEEcompsocitemizethanks{
\IEEEcompsocthanksitem Min Cao and Yucheng Ji are with the School of Computer Science and Technology, Soochow University, SuZhou, China (e-mail: mcao@suda.edu.cn, 20254227067@stu.suda.edu.cn).
\IEEEcompsocthanksitem Mang Ye, Yang Bai, and Bo Du are with the School of Computer Science, Wuhan University, Wuhan, China (e-mail: yemang@whu.edu.cn).

\IEEEcompsocthanksitem Siyuan Chai is with Zhipu AI, Beijing 100089, China.

\IEEEcompsocthanksitem Min Zhang is with Harbin Institute of Technology, Shenzhen, China.

\IEEEcompsocthanksitem Min Cao is the
corresponding author and this work is dominated at Soochow University. 

\IEEEcompsocthanksitem This work was supported by the National Natural
Science Foundation of China under Grants No. 62476188 and T2541022.
}
}


\markboth{IEEE TRANSACTIONS ON PATTERN ANALYSIS AND MACHINE INTELLIGENCE}%
{Shell \MakeLowercase{\textit{et al.}}: Bare Demo of IEEEtran.cls for Computer Society Journals}

\IEEEtitleabstractindextext{%
\begin{abstract}
\justifying
Text-based person search (TBPS) aims to retrieve images of a target person from a large image gallery based on a given natural language description. Most existing methods rely on supervised learning with manually annotated image-text pairs.
In this paper, we explore unsupervised TBPS, with only unlabeled images. 
We propose GTR$^+$, a two-stage generation-then-retrieval framework.
In the generation stage, we introduce a tiered description generation framework designed to produce fine-grained and stylistically diverse textual descriptions through a three-tier sequential process. The base tier leverages an automated question-and-answer mechanism to generate basic visual attribute descriptions; the intermediate tier enhances fine-grained detail using an inter-sample contrastive mechanism; the advanced tier further enriches textual diversity via a stylized expansion mechanism.
In the retrieval stage, to mitigate the impact of noisy pseudo texts, we develop an adaptive confidence-weighted retrieval learning framework. We model image-text pairs as clean or noisy using a Gaussian Mixture Model, calibrated by real-time image-text similarity and static text generation probability from the prior stage, yielding adaptive sample weights during training.
Beyond that, we also contribute LargeFine-Person, a large-scale TBPS dataset with high-quality, fine-grained, and diverse textual annotations, enabling a practical and generalizable TBPS pre-training benchmark under unsupervised setting.
Experiments on multiple TBPS benchmarks demonstrate the effectiveness and generalization of both GTR$^+$ and LargeFine-Person. Code is available at: https://github.com/Flame-Chasers/GTR.

\end{abstract}
}

\maketitle
\IEEEdisplaynontitleabstractindextext

\IEEEpeerreviewmaketitle
\IEEEraisesectionheading{\section{Introduction}\label{sec:intro}}
Text-based person search (TBPS)~\cite{li2025exploring,bai2023rasa,cvpr23crossmodal} involves retrieving images of a specific individual from an image gallery based on textual descriptions. 
This task shares similarities with classical person re-identification~\cite{zhao2021incremental,ye2021deep} and image-text retrieval tasks~\cite{cao2022image,qu2021dynamic}, yet presents unique challenges of its own.
In contrast to person re-identification that relies on image-based queries, TBPS offers a more accessible solution with free-form text as queries, but faces challenges in cross-modal learning due to differences between visual and textual data. 
Compared to general image-text retrieval, TBPS focuses on identifying individuals, which requires attention to fine-grained details and presents greater challenges due to increased intra-class variability (\emph{e.g.,} varying appearances of the same individual) and reduced inter-class variability (\emph{e.g.,} visually similar individuals).
Despite these challenges, TBPS has gained attention in recent years for its potential in real-world applications, such as surveillance for locating suspects or missing persons.

\begin{figure}[t]
  \centering

    \includegraphics[width=\linewidth]{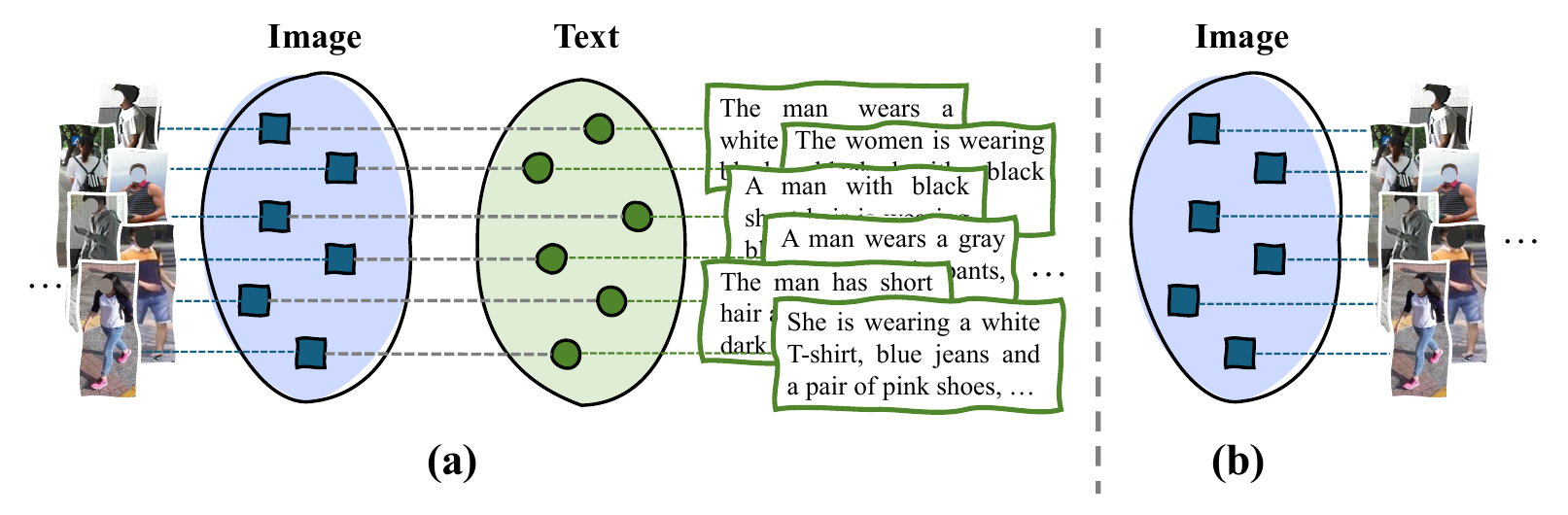}
    \caption{Illustration of (a) canonical TBPS with supervised image-text data, (b) TBPS rely solely on image corpora.}
  \label{fig:intro-fig1}
  \vspace{-0.25cm}
\end{figure}

The current mainstream research in TBPS~\cite{qin2024noisy,li2022learning,bai2025chat} focuses on developing robust modality-invariant representations through sophisticated cross-modal alignment frameworks. While these approaches improve performance, they heavily depend on supervised image-text data (\emph{i.e.,} person images with manually annotated descriptions) for training, as shown in Fig.~\ref{fig:intro-fig1} (a). This reliance presents challenges for real-world applications, where annotating large-scale person images with textual descriptions is labor-intensive and time-consuming.
Therefore, this work focuses on unsupervised TBPS, eliminating the need for annotated text corpora and relying solely on image corpora for effective retrieval, as shown in Fig.~\ref{fig:intro-fig1} (b).
For this, we present a basic two-stage solution, generation-then-retrieval (GTR$^+$), which firstly generates pseudo text descriptions corresponding to the person images for remedying the absent annotation, and then trains a retrieval modal in a supervised manner.


\begin{figure}[t]
  \centering

    \includegraphics[width=\linewidth]{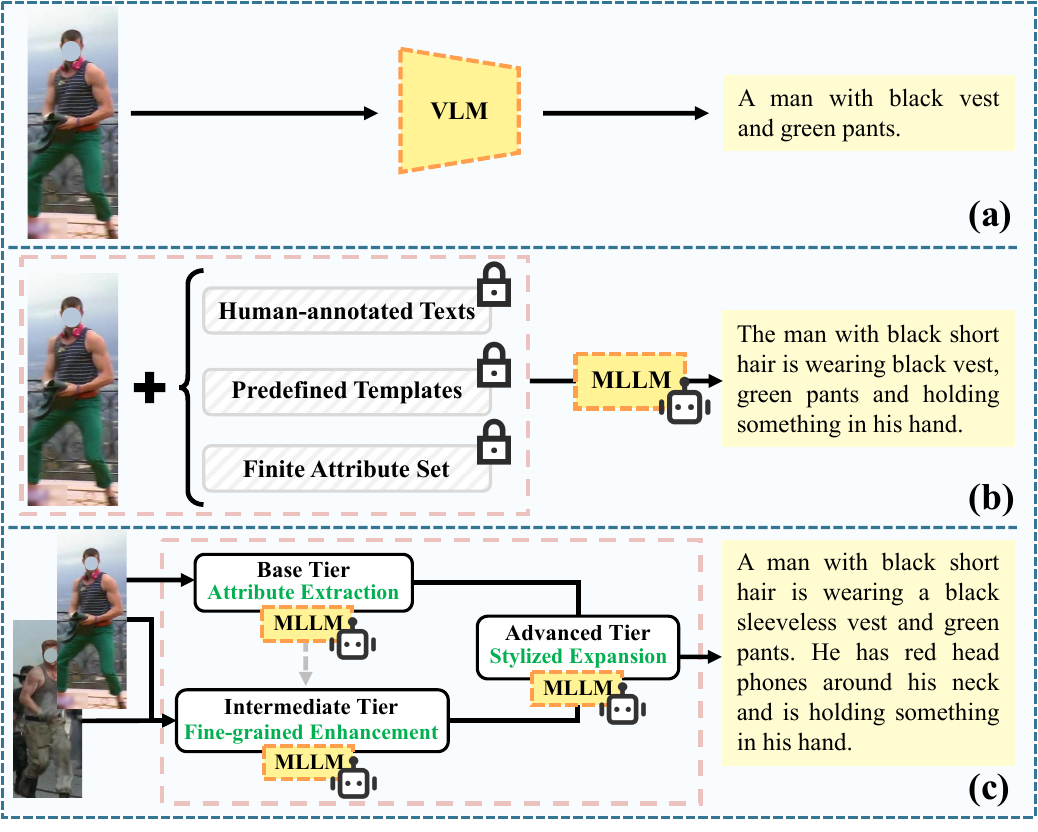}
    \caption{Illustration of the text generation for a given person image. (a) Direct image captioning using a pre-trained vision-language model. (b) Description generation via multimodal large language model with auxiliary information. (c) The proposed tiered description generation framework.}
  \label{fig:intro-fig2}
  \vspace{-0.25cm}
\end{figure}

\textbf{Generation.}
Given the person images, our goal is to generate their paired pseudo textual descriptions that accurately depict the fine-grained details with diverse sentence structures. 
A simple approach is to use pre-trained Vision-Language Models (VLMs)~\cite{li2022blip} for direct image captioning. However, as shown in Fig.~\ref{fig:intro-fig2} (a), these models often fail to produce satisfactory results due to their pre-training on general image-text pairs, limiting their ability to describe the nuanced features specific to person images.
Some studies~\cite{bai2023text,tan2024harnessing,zuo2024plip,jiang2025modeling} leverage Multimodal Large Language Models (MLLMs) for fine-grained description generation. However, these methods often rely on predefined templates, finite attribute set, or annotated texts as guidance. Consequently, the generated textual descriptions provide limited information, restricting both the richness and granularity of the outputs, as shown in Fig.~\ref{fig:intro-fig2} (b).
To address these limitations, we propose a tiered description generation framework which constructs fine-grained and diverse textual descriptions through a self-contained, multi-tier process (Fig.~\ref{fig:intro-fig2} (c)).
Specifically, it consists of three sequential tiers.
1) Base tier. Instead of relying on fixed templates, we design an automated question-and-answer mechanism that enables MLLM to self-query and extract fine-grained visual attributes (\emph{e.g.,} gender, clothing type, clothing color, accessories). This process aims to generate a basic textual description that captures the essential appearance characteristics of the person.
2) Intermediate tier. We develop an inter-sample contrastive mechanism to further capture subtle visual details. By comparing the target image with the hard-negative image, MLLM is prompted to highlight both similarities and differences, enhancing the granularity and informativeness of the descriptions.
3) Advanced tier. Building upon the results of the first two tiers, we propose a stylized expansion mechanism to enhance textual style diversity. 
The mechanism leverages MLLM to automatically enrich descriptions in various styles, guided by the previously generated texts.
Overall, our framework produces fine-grained and stylistically diverse textual annotations in a tier-wise manner, which serve as effective supervision for training TBPS model.

\textbf{Retrieval.}
This stage adopts generated image-text pairs to employ supervised training of the retrieval model. 
The retrieval model can, in principle, be integrated with any existing TBPS method~\cite{cvpr23crossmodal,li2024adaptive}. However, it is important to note that these TBPS methods are typically designed to train on manually annotated and well-aligned image-text pairs. In contrast, the constructed pairs used as training data in our approach may contain noise due to potential hallucinations and inaccuracies introduced during the preceding generation stage.
To mitigate the impact of noise on the retrieval model training, we propose an adaptive confidence-weighted retrieval learning framework. 
Confidence scores, reflecting image-text correspondence reliability, are adaptively incorporated into the loss to calibrate error propagation, assigning higher weights to more reliable pairs and enhancing their influence on learning.
Confidence scores are computed using a Gaussian Mixture Model (GMM) that separates image-text pairs into clean and noisy components based on their similarity distribution. As training progresses, the similarity is updated adaptively—incorporating both real-time image-text alignment and prior text generation probability—enabling the GMM to iteratively refine sample reliability. This progressive refinement sharpens the clean-noisy distinction, yielding a more robust and accurate retrieval model.

In addition to proposing a GTR+ solution for unsupervised TBPS, we also contribute a large-scale TBPS dataset with fine-grained and diverse textual descriptions, named \emph{LargeFine-Person}. 
The dataset is constructed by collecting and refining person images from LUPerson~\cite{fu2021unsupervised} and LPW~\cite{song2018region}, followed by generating corresponding textual descriptions using our proposed tiered description generation framework. As a result, we construct a total of $3.6$ million image-text pairs.
LargeFine-Person enables a practical and generalizable TBPS pre-training benchmark under unsupervised setting.
A comparison between the proposed LargeFine-Person dataset and other pre-training TBPS datasets~\cite{shao2023unified,yang2023towards,tan2024harnessing,zuo2024plip,jiang2025modeling} is presented in Table~\ref{tab:data_comparision}.
LargeFine-Person presents superior data quality—especially in terms of textual descriptions—attributed to the proposed tiered description generation framework.

In summary, our main contributions are as follows:

\begin{enumerate}
\item We propose a tiered description generation framework, integrating automated question-and-answer generation, inter-sample contrastive generation, and stylized expansion generation to produce semantically rich and diverse captions for person images. Based on this, we construct LargeFine-Person, a large-scale dataset of $3.6$ million high-quality, fine-grained image-text pairs.

\item We propose an adaptive confidence-weighted retrieval learning framework that mitigates the impact of noisy pseudo-texts via self-adaptively weighting image-text pairs based on the confidence scores. These scores are iteratively refined using a Gaussian Mixture Model that integrates image-text similarity and text generation probability.
\item We conduct comprehensive experiments on multiple TBPS benchmarks. The results demonstrate the effectiveness and generalizability of our proposed GTR+ solution and LargeFine-Person dataset in unsupervised settings.
\end{enumerate}

A preliminary version of this work has been presented at ACM MM 2023~\cite{bai2023text}. This paper introduces the following significant advancements:
(1) At the task level, while prior work predominantly focuses on TBPS without parallel image-text data—utilizing separately collected image and text corpora where textual data is still required, we focus on the more challenging and practical task of unsupervised TBPS, where only an image corpus is available.
(2) At the model level, we propose a generation-then-retrieval framework, which extends and improves upon previous work in both text generation and retrieval modeling.
In the text generation stage, we design a novel tiered description generation framework, which comprises three progressive components: an automated question-and-answer mechanism to capture basic attributes, an inter-sample contrastive mechanism to encode fine-grained details, and a stylized expansion mechanism to enrich the expressiveness of the generated descriptions. This tiered framework facilitates the generation of descriptions that are both semantically rich and stylistically diverse. It significantly improves upon previous work, which typically relied on manually designed prompts and external text corpora for image-to-attribute extraction and attribute-to-text conversion—often leading to monotonous outputs.
In the retrieval learning stage, we introduce an adaptive confidence-weighted learning framework that enhances robustness to noisy data. Unlike previous work that used a static text generation probability as the confidence score, our method dynamically adjusts confidence scores during training, enabling them to adapt to the learning dynamics. This allows for more accurate weighting of training samples and better noise estimation, ultimately improving retrieval performance.
(3) At the data level, leveraging the proposed tiered description generation framework, we additionally contribute a pre-training dataset LargeFine-Person, which enables a generalizable pre-training benchmark under unsupervised setting. These resources are expected to advance the application of TBPS in real-world scenarios.

\begin{table*}[t]
    \centering
    \resizebox{\linewidth}{!}{
        \setlength\tabcolsep{1.1pt}
        \renewcommand\arraystretch{1.2}

    \begin{tabular}{l||ccC{1.2cm}|C{1.4cm}C{1.4cm}C{1.6cm}C{1.2cm}C{1.2cm}c}
    \hline\thickhline
    \rowcolor{lightgray}
    & \multicolumn{3}{cI}{Image Domain} & \multicolumn{6}{c}{Text Domain}  \\
    \rowcolor{lightgray}
    \multirow{-2}{*}{Datasets}
    & Source & \#Images & Denoise 
    & Source & \#Texts & \#Words/Text & \#Adj. & \#NCs. & Multi-granularity \\
    \hline\hline
    MALS~\cite{yang2023towards} & Generation &  1.51M  & \cmark 
    & Human. & 1.51M  & 27 & 2.72 & 7.97 & \xmark \\
    LUPerson-T~\cite{shao2023unified}
    & LUPerson~\cite{fu2021unsupervised} & 1.30M & \xmark 
    & Auto. & 1.30M  & 27 & 4.39 & 5.86 & \xmark \\
    LUPerson-MLLM~\cite{tan2024harnessing} & LUPerson~\cite{fu2021unsupervised}  &  1.00M  & \xmark 
    & Auto. & 4.00M  & 29 & 3.47 & 8.62 & \xmark \\
    SYNTH-PEDES~\cite{zuo2024plip} &  LUPerson-NL~\cite{fu2022large} + LPW~\cite{song2018region} & 4.79M & \cmark 
    & Human. & 12.14M  & 24 & 4.81 & 7.08 & \xmark \\
    HAM-PEDES~\cite{jiang2025modeling} & SYNTH-PEDES~\cite{zuo2024plip} &  1.00M  & \xmark 
    & Human. & 2.00M  & 26 & 5.66 & 7.54 & \xmark \\
    \hline 
    \rowcolor[HTML]{D7F6FF}
    LargeFine-Person (Ours) & LUPerson~\cite{fu2021unsupervised} + LPW~\cite{song2018region} & 1.20M  & \cmark 
    & Auto. & 3.60M  & 61 & 9.63 & 15.27 & \cmark \\
    \hline\thickhline
    \end{tabular}
    }
    \caption{Comparison between our pre-training dataset (LargeFine-Person) and existing pre-training datasets for TBPS. 'Denoise' indicates whether low-quality images are filtered out from the dataset. 'Human.' specifies that the generation of texts depends on manually annotated inputs, while 'Auto.' denotes that the text generation is fully automated without manual intervention. '\#Words/Text' represents the average number of words per text, reflecting the richness of information conveyed. '\#Adj.' denotes the average number of adjectives per text, reflecting the fine-grainedness of the descriptions. '\#NCs.' denotes the average number of noun chunks per text, indicating the completeness of the attribute coverage. 'Multi-granularity' indicates the presence of texts with varying levels of detail, reflecting the diversity of the textual data. Benefiting from the proposed tiered description generation framework, our dataset demonstrates advantages in both diversity and fine-grained detail.}
    \label{tab:data_comparision}
    \vspace{-0.25cm}
\end{table*}

\section{Related Work}
\subsection{Text-based person search}

Existing TBPS approaches can be categorized into two main paradigms based on their primary focus: cross-modal alignment and representation learning.
The first line of research focuses on cross-modal alignment, aiming to map visual and textual features into a shared embedding space. Early methods relied on global feature alignment~\cite{li2017person,zheng2020dual, farooq2020convolutional}, while later works progressed to multi-granularity correspondences~\cite{chen2022tipcb, suo2022simple, wu2024laip, qi2025granularity, li2024adaptive} and incorporated external tools—such as human parsing, pose estimation, and NLTK~\cite{loper2002nltk}—for fine-grained feature alignment~\cite{wang2020vitaa, jing2020pose, zheng2020hierarchical}. Recent advances have achieved progress in self-adaptive semantic alignment~\cite{li2022learning, gao2021contextual, gao2022conditional, ji2022asymmetric}.
The second paradigm focuses on learning modality-invariant representations that generalize across modalities~\cite{shao2022learning, yan2024prototypical, lu2025prompt, park2024plot,yan2024prototypical}. For instance, Wu \emph{et al.}~\cite{wu2021lapscore} introduced color-reasoning sub-tasks to enhance fine-grained cross-modal associations, while Wang \emph{et al.}~\cite{wang2022caibc} proposed a color deprivation module to reduce over-reliance on color and learn more robust features.
Recent efforts~\cite{cvpr23crossmodal, han2021text, yan2022clip, wei2023calibrating, cao2024empirical, lu2025prompt, bai2023rasa} have utilized large vision-language models like CLIP~\cite{radford2021learning} and multimodal LLMs to enhance fine-grained retrieval. 
Several studies~\cite{qin2025human, bai2025chat} further incorporated multi-round interactions and contextual scene information~\cite{xu2025sa} for better query-image alignment.
However, these methods still rely heavily on costly, labor-intensive manually annotated image-text pairs, limiting their scalability and real-world applicability.

Recently, researchers have begun to recognize the annotation burden inherent in traditional TBPS and have explored less annotation-intensive paradigms. 
Several studies~\cite{zhao2021weakly, zheng2024cpcl, fu2025similarity, gong2024enhancing, zhang2025dual} focused on weakly supervised TBPS, where image-text pairs are available without explicit identity labels. 
Gao et al.~\cite{gao2024semi} explored a semi-supervised setting with limited labeled pairs and abundant unlabeled images, while Jing et al.~\cite{jing2020cross} studied cross-domain adaptation from labeled source domains to unlabeled targets. 
Though reducing annotation effort, these methods still require partial supervision.
In contrast, fully unsupervised TBPS methods~\cite{li2024cross, bai2023text, li2025exploring} required only unannotated images. They typically generated pseudo-text annotations followed by noise-robust learning. However, the performance is limited by: (1) template-based text generation yields limited and single-perspective descriptions, and (2) existing noise modeling strategies consider only local text-image similarity.
In this work, we solve unsupervised TBPS that addresses both limitations via three-tiered text generation and adaptive confidence-weighted retrieval learning.

\subsection{Unsupervised Vision-Language Tasks}
Considering that the paired image-text data are extremely labor-intensive to collect, significant efforts have been dedicated to advancing unsupervised learning in vision-language tasks, including unsupervised vision-language pretraining~\cite{li2021unsupervised, zhou2022unsupervised, chen2022end, wang2022vlmixer}, unsupervised text-to-image synthesis~\cite{dong2021unsupervised}, and unsupervised image captioning~\cite{feng2019unsupervised, gu2019unpaired, guo2021recurrent, ben2021unpaired,huang2024unpaired}.
In particular, unsupervised image captioning has seen flourishing development, driven by two major settings: unpaired image captioning (training with image-text data but unpaired each other)~\cite{zhu2023prompt,huang2024unpaired,huang2022mack} and text-only image captioning (training exclusively on text corpora)~\cite{qi2024relational,zeng2024meacap,li2023decap}.
For text-only image captioning, some studies~\cite{feng2019unsupervised,qi2024relational,zeng2024meacap} explored leveraging visual concepts to bridge the gap between textual and visual modalities, while others~\cite{liu2025synthesize,luo2024unleashing,liu2024improving} proposed generating synthetic images from text data, training captioning models using these synthetic image-text pairs. Additionally, some research~\cite{qiu2024mining,li2023decap,fei2023transferable} focused on exploiting the cross-modal alignment capabilities of vision-language pretraining models by designing auxiliary modules to better adapt these models to downstream image captioning task.
Nevertheless, these advancements in unsupervised vision-language tasks cannot be directly applied to unsupervised TBPS due to the significant differences in tasks. 
Conventional unsupervised vision-language tasks focus on broad, universal semantic content, often relying on unpaired image-text or text-only corpora. 
In contrast, unsupervised TBPS emphasizes fine-grained, individual-specific categories using image-only corpora, posing compatibility challenges.

\subsection{Image Captioning}
Image captioning~\cite{yang2021deconfounded} is a vision-language task that aims to generate accurate and fluent natural language descriptions of images. Early approaches~\cite{yao2010i2t} relied on simple template-based sentence generation, dependent on object detectors or attribute predictors. 
Subsequent research~\cite{yang2020auto,vinyals2015show} transitioned to leveraging Convolutional Neural Networks (CNNs) to encode visual features, which were then decoded into coherent textual descriptions using Recurrent Neural Networks (RNNs).
Recently, Transformer-based architectures~\cite{cornia2022explaining,al2025ensemble,barraco2023little,hessel2021clipscore} have emerged as the dominant approach, being applied both in the visual encoding stage and in the language modeling process. Furthermore, the rapid advancements in multimodal large language models (MLLMs)~\cite{touvron2023llama,chen2024sharegpt4v,chen2025blip3} have introduced new paradigms for image captioning~\cite{yang2023exploring,bucciarelli2025personalizing,dong2024benchmarking}. These models utilized pre-trained vision-language representations to generate more coherent, semantically meaningful captions, significantly advancing the integration of vision and language understanding.
Person images in TBPS differ from image captioning, focusing on fine-grained, individual-specific features rather than general semantics. Existing image captioning methods are unsuitable for TBPS. We propose a tired generation framework to enable fine-grained textual descriptions.

\subsection{Cross-modal Noise-robust Learning} 
The problem of noisy correspondence in cross-modal learning has garnered increasing attention recently.
Huang \emph{et al.}~\cite{huang2021learning} first identified this issue and introduced a noisy correspondence rectifier that employs neural memorization techniques to categorize data into clean and noisy subsets, subsequently correcting misalignments adaptively. 
Building on this, Han \emph{et al.}~\cite{han2023noisy} proposed a meta-learning purification strategy aimed at suppressing noisy samples, Dang \emph{et al.}~\cite{dang2025disentangled} developed an information-theoretic framework designed to enhance feature disentanglement in noisy correspondence scenarios.
More recently, Shi \emph{et al.}~\cite{shi2024breaking} presented a general framework for filtering out noisy correspondences by utilizing priors from pretrained models. 
Beyond that, in the TBPS domain, several studies have also addressed the challenge of noisy correspondences.
Qin et al.~\cite{qin2024noisy} handled misaligned person image-text pairs through a confident consensus mechanism for noise filtering and a triplet alignment loss to avoid misleading supervision during training.
Another tackled the issue of noise originating from weakly matched image-text pairs, where parts of the text are irrelevant to the corresponding image, often due to pairing images with generated pseudo-texts. 
Gao et al.~\cite{gao2024semi} proposed a hybrid patch-channel masking alongside a noise-guided progressive training strategy to boost model robustness against noise. 
Tan et al.~\cite{tan2024harnessing} introduced a noise-aware masking technique to identify and mask incorrect words based on token similarity. 
Li et al.~\cite{li2025exploring} further contributed to this field by developing an uncertainty-guided filtration module using a Gaussian mixture model to identify and filter noisy samples.
The earlier version of this work~\cite{bai2023text} explored confidence score-based training to prioritize reliable samples based on fixed confidence scores derived during the generation process. Expanding on this concept, we further develop an adaptive confidence-weighted retrieval learning framework to self-adaptively weight samples.

\begin{figure*}[t]
  \centering
  \setlength{\abovecaptionskip}{4pt}
  \setlength{\belowcaptionskip}{-10pt}
  
  \includegraphics[width=0.95\linewidth]{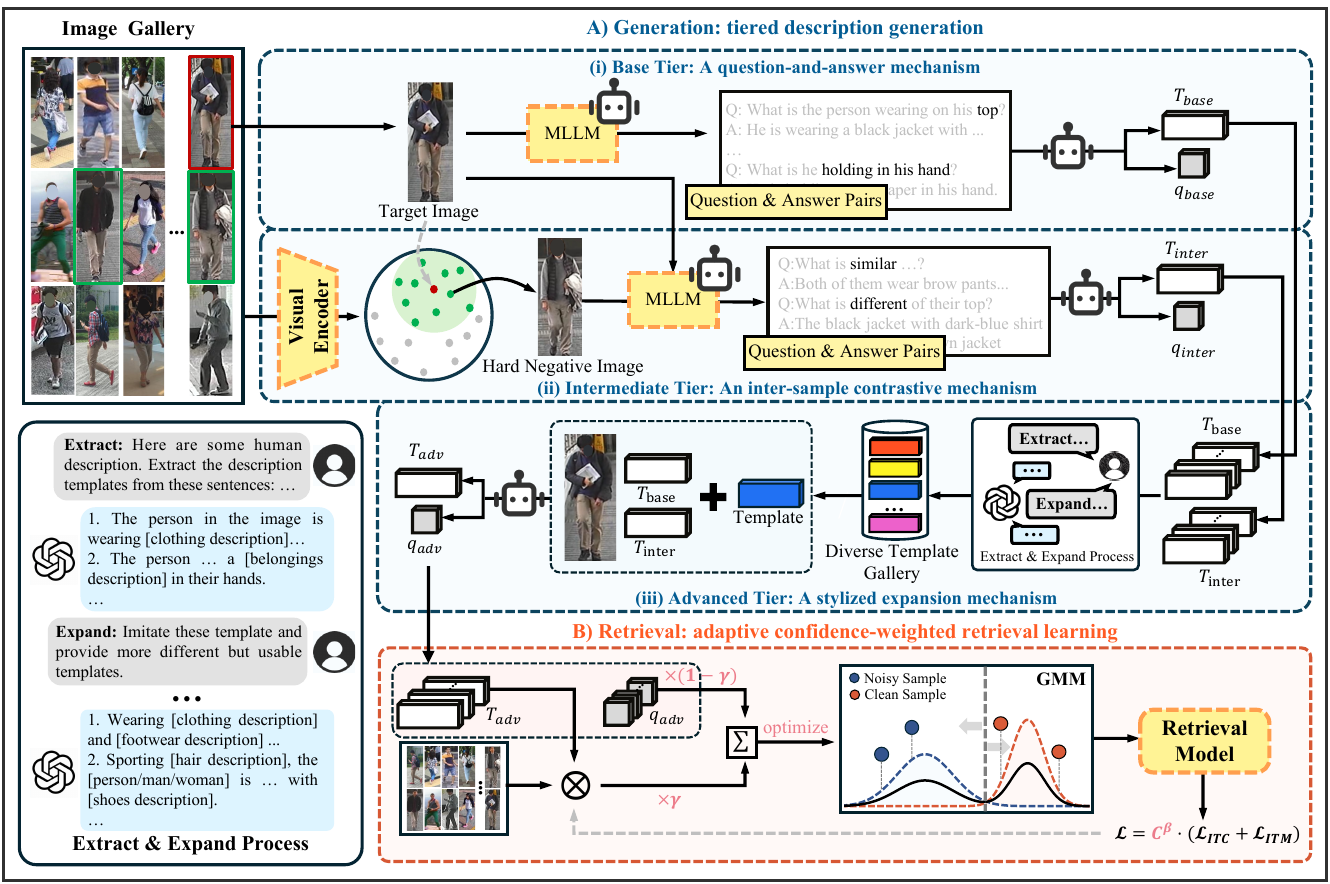}
  \caption{The overview of the proposed generation-then-retrieval solution (GTR+). The proposed solution consists of a tiered description generation framework and an adaptive confidence-weighted retrieval learning framework. In the first stage, the tiered description generation framework generates fine-grained and diverse textual descriptions through three hierarchical tiers. (i) Base Tier: a question-and-answer mechanism is introduced to generate basic descriptive content; (ii) Intermediate Tier: an inter-sample contrastive mechanism enhances descriptions with finer details; (iii) Advanced Tier: a stylized expansion mechanism diversifies the linguistic styles of the generated text.  
  In the second stage, the adaptive confidence-weighted retrieval learning framework enables noise-robust training by mitigating the impact of noisy textual descriptions generated in the previous stage.}
  \label{fig:architecture}
\end{figure*}

\section{Method}


We focus on solving unsupervised TBPS, in which  only a collection of person images $\mathbb{S}_I = \{I_1, I_2, \cdots\}$ is provided. 
In the section, we formally delineate the proposed two-stage solution, generation-then-retrieval (as illustrated in Figure \ref{fig:architecture}).
The solution consists of a novel tiered description generation framework (detailed in Section~\ref{subsec:generation}) that automatically generates synthetic text data $\mathbb{S}_{T^{syn}} = \{T^{syn}_1, T^{syn}_2, \cdots\}$, and an adaptive confidence-weighted retrieval learning (detailed in Section~\ref{subsec:retrieval}) that enables noise-robust training of the retrieval model based on both the original image data $\mathbb{S}_I$ and the synthetic text data $\mathbb{S}_{T^{syn}}$.
For simplicity, we next omit symbolic superscripts and subscripts, using $I$ and $T$ to represent an image and a text, respectively.

\subsection{Tiered Description Generation}
\label{subsec:generation}
\noindent
In unsupervised TBPS, there are no annotated text data paired with image data, making the first stage focus on generating pseudo texts for each person image.
However, pre-trained vision-language models~\cite{li2022blip} fall short in capturing fine-grained, person-specific details, and template-based methods~\cite{li2024cross,li2025exploring,zuo2024plip,bai2023text} produce limited, single-perspective descriptions. 
To get an enriched description, we propose a three-tiered description generation framework.

\subsubsection{Base Tier: Question-and-Answer Mechanism}

In this tier, we target to generate a basic textual description that captures the essential appearance attributes of the person. Rather than the previous version of this work~\cite{bai2023text} that rely on template-based generation, we leverage an MLLM to generate the basic textual description through a self-question-and-answer approach.



\begin{figure}[t]
  \centering

    \includegraphics[width=\linewidth]{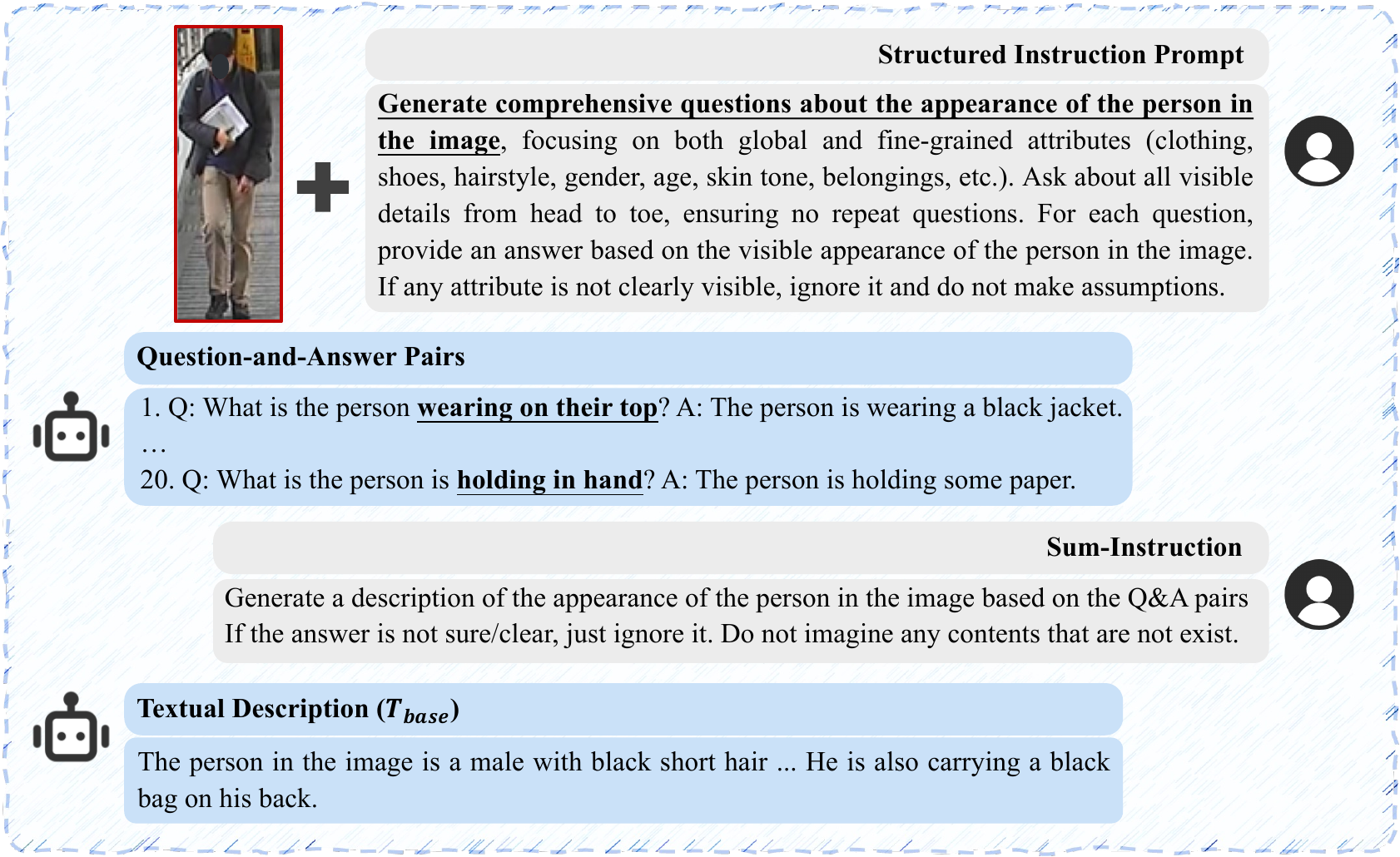}
    \caption{Flowchart of the question-and-answer mechanism.}
  \label{fig:method-fig1}
  \vspace{-0.25cm}
\end{figure}

Specifically, as illustrated in Figure \ref{fig:method-fig1}, given an input image, we design a structured instruction prompt to guide the MLLM in generating a set of question-and-answer pairs. These pairs probe explicit attribute information of the person, covering global characteristics (\emph{e.g.,} gender, age) and fine-grained details (\emph{e.g.,} shoe color, accessories). This approach ensures the model performs exhaustive yet grounded attribute queries and responses without requiring manual intervention.
Subsequently, we apply a sum-instruction to guide the MLLM in extracting and summarizing the information from these question-and-answer pairs, thereby generating a basic textual description that naturally captures the key visual attributes of the image.

This mechanism enables an end-to-end automated approach for extracting pedestrian attributes and generating textual descriptions, eliminating the need for manual intervention and significantly enhancing generalization across diverse images.



\begin{figure}[t]
  \centering

    \includegraphics[width=\linewidth]{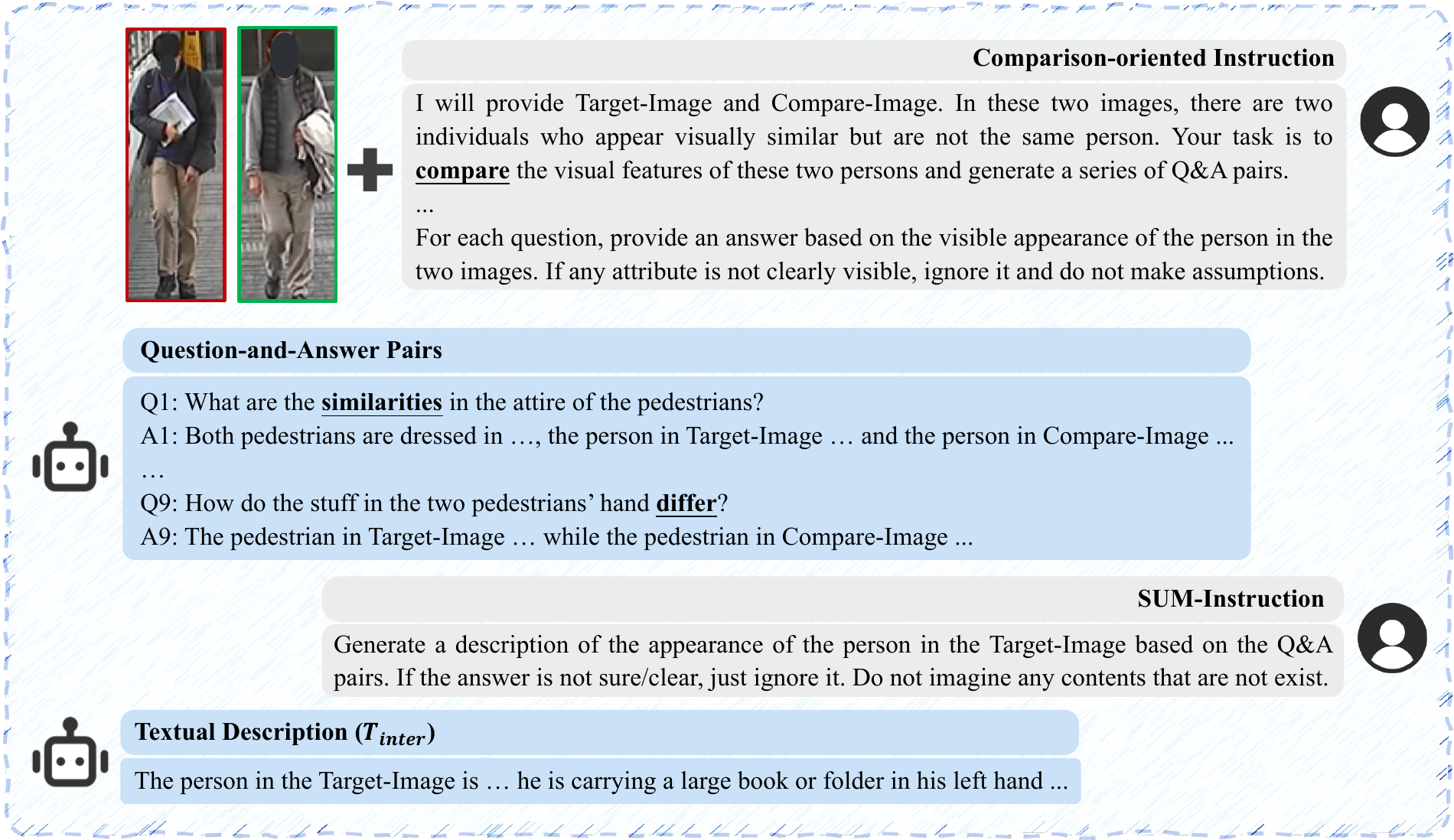}
    \caption{Flowchart of the inter-sample contrastive mechanism.}
  \label{fig:method-fig2}
\end{figure}

\subsubsection{Intermediate Tier: Inter-Sample Contrastive Mechanism}
The base tier generates descriptions containing only basic attribute information, and the intermediate tier improves both granularity and informativeness by incorporating hard negative samples.
The MLLM is prompted to compare the target image with the hard negative image, emphasizing similarities and differences. This process enriches the description with finer details and deeper contextual information, improving its quality.

Specifically, we first utilize the DINO-v2~\cite{oquab2023dinov2} to extract the visual feature $f^{img}$ of each image in $\mathbb{S}_I$. Next, we compute pairwise similarity between images using cosine similarity,
\begin{equation}
s{(f^{img}_{i}, f^{img}_{j})} = \frac{f^{img}_i \cdot f^{img}_j}{||f^{img}_i|| \cdot ||f^{img}_j||},
\end{equation}
where $f^{img}_{i}$ and $f^{img}_{i}$ denote the feature of $i$-th and $j$-th ($i \neq j$) images, repsectively.
For the target image, we identify its top-$10$ nearest neighbors based on pairwise similarity and randomly select one as a hard negative image. 
Instead of always selecting the top-1 nearest neighbor as the hard negative, the random selection from the top-$10$ nearest neighbors ensures greater diversity in the negative samples.
The picked hard negative, while visually similar to the target, is semantically distinct, providing a challenging yet informative contrastive example.

Given the target image and its paired hard negative image, as shown in Figure \ref{fig:method-fig2}, we enhance the instruction prompt from the base tier by introducing a comparison-oriented instruction. This enhanced instruction guides the MLLM to generate a set of question-and-answer pairs that highlight both shared visual elements (\emph{e.g.,} 'both subjects wear striped shirts') and distinctive visual features (\emph{e.g.,} 'distinct shoe colors: white vs. black') between two images.
Subsequently, we apply a revised SUM-Instruction, specifically tailored for the comparison phase, to the MLLM for generating the final text description. This revised SUM-Instruction ensures that the generated caption remains focused solely on the target image while being grounded in the comparative context provided by the hard negative image.

\subsubsection{Advanced Tier: Stylized Expansion Mechanism}\label{subsubsection: adv tier}
Through the aforementioned two tiers, we obtain textual descriptions that capture essential appearance attributes and fine-grained details of the person images. However, we observe that MLLMs tend to generate captions with limited sentence structures when provided with identical instruction prompts. This structural uniformity in the descriptions could result in overfitting to specific linguistic patterns, thereby diminishing the model's ability to generalize effectively in real-world applications.
To address this, we consider to explicitly diversify the styles of textual descriptions in this tier.


Specifically, we begin by randomly sampling a subset of descriptions obtained from the aforementioned two tiers and using ChatGPT-4o to analyze their underlying syntactic templates. Through multiple rounds of interaction with ChatGPT-4o, we iteratively refine these templates to create a diverse array of sentence patterns, which are incorporated into a diverse template gallery.

During textual description generation, a template is randomly selected from the diverse template gallery. The MLLM is then prompted to follow the chosen template and generate a detailed textual description (denoted as $T_{adv}$) using the following instruction,
\begin{tcolorbox}
\textit{Generate a detailed description of the person's overall appearance based on the {$T_{base}$} and {$T_{inter}$}, using the following template: \{template\}. Do not invent or assume any details that are not evident in the image.}
\end{tcolorbox}
Here, $T_{base}$ and $T_{inter}$ denote the texts generated by the base tier and the intermediate tier, respectively, and \{template\} is placeholder that is replaced with the chosen template. 


Finally, for each image, $T_{adv}$ is generated as the paired text. Since $T_{adv}$ already incorporates descriptive information from both $T_{base}$ and $T_{inter}$, we use only $T_{adv}$ in constructing the final image-text pair.
To further enhance the diversity of the generated texts, we leverage two state-of-the-art, publicly available MLLMs: Qwen2-VL-7B-Instruct~\cite{wang2024qwen2} and InternVL2.5-8B~\cite{chen2024expanding} for generation, respectively. As a result, each image is paired with two texts, forming the final training dataset. 




\subsection{Adaptive Confidence-Weighted Retrieval Learning}
\label{subsec:retrieval}
\noindent

The generative stage produces synthetic image-text pairs that, in principle, could serve as supervised signals for training the retrieval model. 
However, the inherent propensity of MLLMs to introduce hallucinations and inaccuracies during generation leads to noise, thus hindering downstream retrieval performance.
To mitigate the effects of such noise, we develop an adaptive confidence-weighted retrieval learning framework.

\subsubsection{Baseline}
We first adopt BLIP~\cite{li2022blip} as the baseline retrieval model, and yet it can be replaced with others (as verified experimentally in Section~\ref{sec:Dif-baseline}) due to the flexible nature of the proposed retrieval learning framework.

BLIP consists of an image encoder $\mathbf{E}_{img}$, a text encoder $\mathbf{E}_{txt}$, and an image-grounded text encoder $\mathbf{E}_{i\&t}$.
Given an image $I=\{ p_1, p_2, \cdots\}$ and a generated text $T=\{ w_1, w_2, \cdots \}$, where $p_i$ is $i$-th non-overlapping patch and $w_j$ is $j$-th token, we compute the global feature of image, the global feature of text, and the multimodal feature of the image-text pair as,
\begin{equation}
f^{img} = \mathbf{E}_{img}(I), \quad f^{txt} = \mathbf{E}_{txt}(T),
\end{equation}
\begin{equation}
f^{i\&t} = \mathbf{E}_{i\&t}(\mathbf{E}_{img}(I), \mathbf{E}_{txt}(T)).
\end{equation}

Then, the optimization objectives with an image-text contrastive learning (ITC) loss and an image-text matching (ITM) loss are computed as,
\begin{equation}
\begin{gathered}
L_{itc}^{i2t} = -E_{(I,T)\sim D} \log{\frac{\exp(s(f^{img},f^{txt})/\tau)}{\sum_{m=1}^M \exp(s(f^{img},f^{txt}_m)/\tau)}}, \\
L_{itc}^{t2i} = -E_{(I,T)\sim D} \log{\frac{\exp(s(f^{img},f^{txt})/\tau)}{\sum_{m=1}^M \exp(s(f^{img}_m,f^{txt})/\tau)}}, \\
L_{itc} = (L_{itc}^{i2t} + L_{itc}^{t2i}) \ / \ {2},
\end{gathered}
\label{eq_4}
\end{equation}
where $M$ is the number of instances in a mini-batch, $s(f^{img},f^{txt})$ measures the cosine similarity between the image and the text, $\tau$ is a learnable temperature;

\begin{equation}
L_{itm} = \mathbb{E}_{(I,T)\sim D} \mathcal{H}(y, \mathcal{\phi}(f^{i\&t})),
\label{eq_5}
\end{equation}
where $\mathcal{H}$ denotes the cross-entropy function, and $y$ is 2-dimension one-hot vector and denotes the ground-truth label. $\mathcal{\phi}(f^{i\&t})$ is the predicted matching probability of the pair and $\mathcal{\phi}(\cdot)$ is a linear classification layer.


\subsubsection{Adaptive Confidence Weighting for Noise-Robust Retrieval}
In the baseline retrieval model, the training objectives (Eq.~\ref{eq_4} and Eq.~\ref{eq_5}) treat all image-text pairs equally, assigning uniform weights during optimization. 
However, noise in the generated texts can mislead training and degrade performance. To address this, we introduce adaptive confidence weighting into the objectives, assigning higher weights to clean samples and lower weights to noisy ones, thereby mitigating the impact of noise.

The confidence score reflects the likelihood of the pair being clean, with high value indicating clean data and vice versa. To estimate the score, we model the similarity distribution of image-text pairs in each training epoch. Based on this distribution, we compute the probability that each pair belongs to the clean data distribution, which is then used as the confidence score for weighting in the subsequent training step.


Specifically, we employ the Gaussian Mixture Model (GMM), which is well known for its ability to capture complex distributions by combining multiple Gaussian components, to model the similarity distribution of image-text pairs,
\begin{equation}
p(\hat{s}) = \sum_{k=1}^{K} \alpha_k \, \mathcal{N}(\hat{s} \mid \mu_k, \sigma_k^2).
\label{eq-GMM}
\end{equation}
Here, we set $K=2$, corresponding to two underlying distributions: one representing clean data and the other representing noisy data.
The parameter $\alpha_k$ denotes the mixing weight of the $k$-th Gaussian component, satisfying $\alpha_1 + \alpha_2 = 1$ and $\alpha_k \geq 0$ for all $k$. 
The function $\mathcal{N}(\hat{s} \mid \mu_k, \sigma_k^2)$ denotes the probability density function of a univariate Gaussian, where $\mu_k$ and $\sigma_k^2$ are the mean and variance of the 
$k$-th component, respectively.
And $\hat{s}$ represents the positive pair similarity and is computed as,
\begin{equation}
\hat{s}(f^{img},f^{txt})=\gamma \cdot s(f^{img},f^{txt})+ (1-\gamma)\cdot  q^{txt},
\label{eq-sum}
\end{equation}
where the cosine similarity $s(f^{img},f^{txt})$ is rectified by integrating the text generation probability $q^{txt}$ in $\gamma$-weighted manner.
Specifically, during the previous tiered description generation stage, token-level generation probabilities can outputted from the MLLM. Based on them, the text generation probability $q^{txt}$ is calculated as the mathematical expectation of the generation probabilities of its constituent tokens, serving as a measure of central tendency:
\begin{equation}
q^{txt} = \sum_i^{N}\frac{1}{N} q^{tok}_i,
\end{equation}
where $q^{tok}_i$ is $i$-th token generation probability and $N$ is the number of tokens composing the text. 
A lower value of $q^{txt}$ indicates higher uncertainty in the correspondence between the generated text and the associated image, implying potential noise.
Notably, the previous version of this work~\cite{bai2023text} computed the confidence score as the joint probability of token-level scores, which often results in vanishing values for longer captions and thus limited effectiveness. In contrast, our expectation-based confidence measure offers a more objective and robust assessment of text confidence.

The Gaussian Mixture Model in Eq.~\ref{eq-GMM} is optimized using the Expectation-Maximization (EM) algorithm, based on which we compute the posterior probability that a positive pair belongs to the clean data distribution as its confidence score,
\begin{equation}
C = \frac{\alpha_{k^*} \, \mathcal{N}(\hat{s} \mid \mu_{k^*}, \sigma_{k^*}^2)}{\sum_{k=1}^{2} \alpha_k \, \mathcal{N}(\hat{s} \mid \mu_k, \sigma_k^2)}, \quad k^* = \arg\max_{k} \mu_k.
\end{equation}
Here, the component with the larger mean, $\mu_{k^*}$, represents the clean data distribution.

Finally, we incorporate the confidence score into the optimization objectives as,
\begin{equation}
\begin{gathered}
L_{itc}^{i2t} = -E_{(I,T)\sim D} C^\beta\log{\frac{\exp(s(f^{img},{f}^{txt})/\tau)}{\sum_{m=1}^M \exp(s(f^{img},{f}^{txt}_m)/\tau)}}, \\
L_{itc}^{t2i} = -E_{(I,T)\sim D} C^\beta\log{\frac{\exp(s(f^{img},{f}^{txt})/\tau)}{\sum_{m=1}^M \exp(s(f^{img}_m,{f}^{txt})/\tau)}}, \\
L_{itc} = (L_{itc}^{i2t} + L_{itc}^{t2i}) \ / \ {2},
\end{gathered}
\label{eq:loss_itc}
\end{equation}
\begin{equation}
L_{itm} = \mathbb{E}_{(I,T)\sim D} C^\beta \mathcal{H}(y, \mathcal{\phi}(\hat{f}^{i\&t})),
\label{eq:loss_itm}
\end{equation}
where $\beta$ is a hyper-parameter to control the importance of the confidence score.

During each training batch, the model with confidence-weighted objectives updates the similarity distributions and returns the dynamically updated parameters of the GMM, allowing confidence scores to adapt accordingly. These scores are then incorporated into the training objectives.
This alternating iterative process ensures that confidence estimation remains sensitive to the current training dynamics, yielding more accurate assessments of data noise compared to using a fixed text generation probability $q^{txt}$ as the confidence score—as done in prior work~\cite{bai2023text}.
Moreover, due to Eq.~\ref{eq-sum}, $q^{txt}$ provides relatively reliable guidance in the early stages of training, when the model is still unstable and the cosine similarity $s(f^{img},f^{txt})$ may be unreliable. As training progresses and features become more discriminative, the cosine similarity increasingly captures meaningful semantic relationships. By jointly modeling both signals within a unified framework, the $\beta$-weighted addition ensures robustness to noisy data across different training phases.

\subsection{Pre-training Dataset Construction: LargeFine-Person}

\begin{figure*}[t]
  \centering

  \includegraphics[width=0.94\linewidth, height=0.55\linewidth]{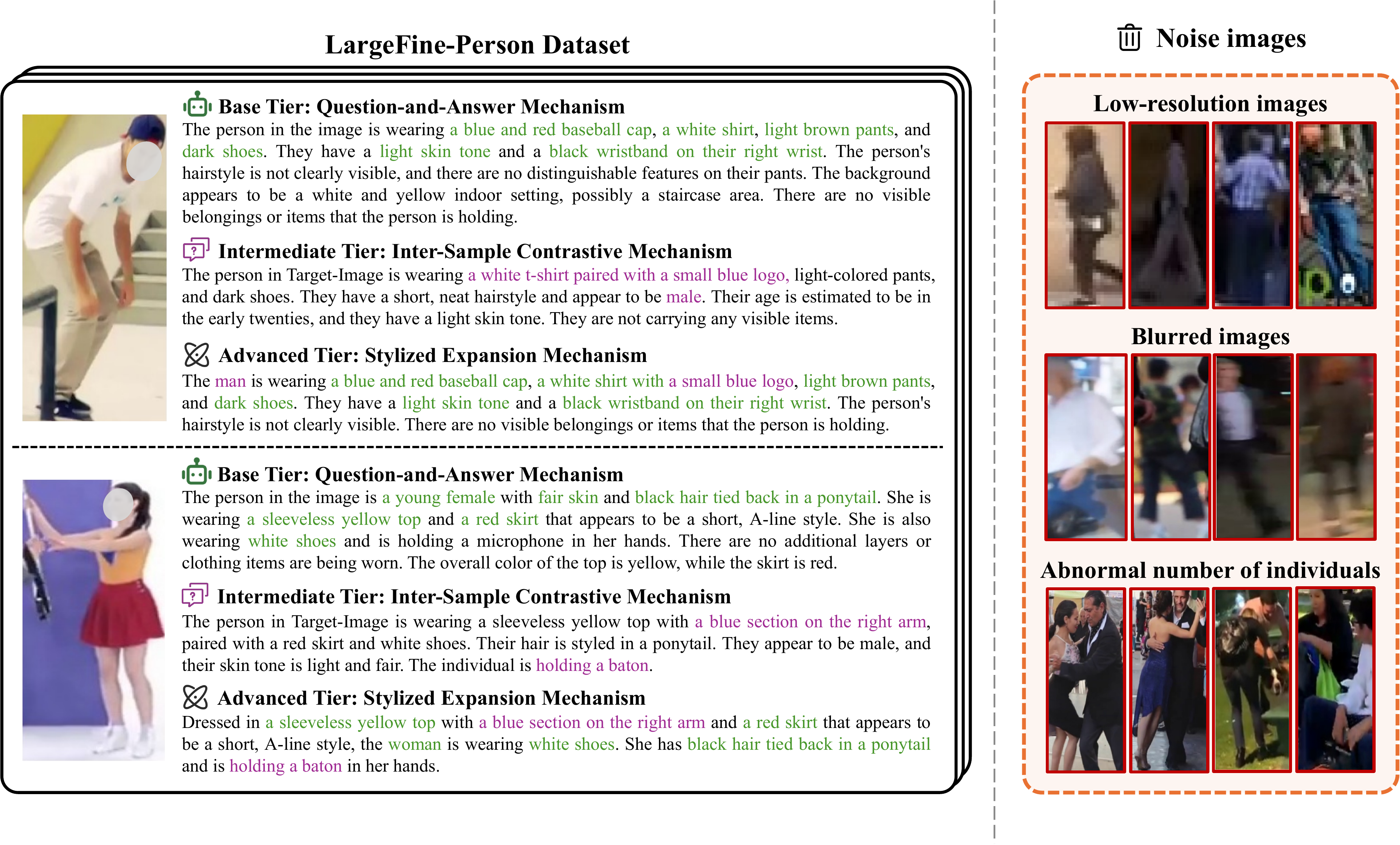}
  \caption{Illustration of the proposed LargeFine-Person dataset and the screened noise images during dataset construction. To ensure data diversity and scale, we generate three textual descriptions for each image, corresponding to the three tiers of our tiered description generation framework, which capture different aspects of person appearance and attributes.}
  \label{fig:dataset}
\end{figure*}

In addition, leveraging the proposed tiered description generation framework, we introduce LargeFine-Person, a large-scale and fine-grained TBPS dataset. 
It is specifically designed to serve as a generalizable pre-training benchmark for TBPS in unsupervised setting.

\textbf{Image Collection.}
We adopt LUPerson~\cite{fu2021unsupervised} and LPW~\cite{song2018region} datasets as the image sources. LUPerson contains $4$ million (M) person images cropped from YouTube videos, while LPW includes $0.59$M person images captured by multi-camera systems. Together, these datasets provide a rich variety of environments.
However, since the images in LUPerson are extracted from videos using algorithms, they inevitably contain a substantial amount of noise. Thus, we perform data denoising on LUPerson to filter out low-quality images based on the following criteria:
\begin{itemize}
    \item Low-resolution images: Images with a width smaller than 32 pixels or a height smaller than 64 pixels are filtered out.
    \item Blurred images: We use the variance of the Laplacian response to measure image sharpness. Images with a Laplacian variance smaller than 80 are regarded as blurry and filtered out.
    \item Abnormal person instances: We utilize YOLOv8 for person detection. Images with no detection box or more than one detection box are filtered out. 
\end{itemize}

As a result, we obtain $0.61$M high-quality images from the LUPerson dataset, which, combined with the $0.59$M images from the LPW dataset, yield a total of $1.2$M images.

\textbf{Text Generation.}
Using the tiered description generation framework, we generate textual descriptions for each of the $1.2$M images, incorporating all generated texts ($T_{base}$, $T_{inter}$, and $T_{adv}$) to enhance diversity. To maintain generation efficiency, we randomly utilize either Qwen2-VL-7B-Instruct~\cite{wang2024qwen2} or InternVL2.5-8B~\cite{chen2024expanding} within this framework. 
Finally, this process yields three textual descriptions per image, resulting in a total of $3.6$M image-text pairs.

In summary, the proposed tiered description generation framework enables the construction of LargeFine-Person entirely without manual annotation. The characteristics of LargeFine-Person are summarized in Table~\ref{tab:data_comparision}, and example illustrations are shown in Figure~\ref{fig:dataset}.

\begin{table*}[t]
    \centering
    \resizebox{0.98\linewidth}{!}{
        \renewcommand\arraystretch{1.2}

    \begin{tabular}{l|l||ccccIccccIcccc}
    \hline\thickhline
    \rowcolor{lightgray}
    & & \multicolumn{4}{cI}{CUHK-PEDES} & \multicolumn{4}{cI}{ICFG-PEDES} & \multicolumn{4}{c}{RSTPReid}  \\

    \rowcolor{lightgray}
    \multirow{-2}{*}{Methods}
    & \multirow{-2}{*}{Reference} & R@1 & R@5 & R@10 & mAP & R@1 & R@5 & R@10 & mAP & R@1 & R@5 & R@10 & mAP \\
    \hline\hline
    \multicolumn{14}{l}{\textbf{\textit{Supervised TBPS}}} \\
    \hline
    Dual Path~\cite{zheng2020dual} 
    & TOMM-20 & 44.40 & 66.26 & 75.07 & - & 38.99 & 59.44 & 68.41 & - & - & - & - & - \\
    ViTAA~\cite{wang2020vitaa}
    & ECCV-20 & 55.97 & 75.84 & 83.52 & 51.60 & 50.98 & 68.79 & 75.78 & - & - & - & - & - \\
    CMKA~\cite{niu2020improving} & TIP-20 & 54.69 & 73.65 &  81.86 & - & 46.49 & 67.14 & 75.18 & - & - & - & - & - \\
    LapsCore~\cite{wu2021lapscore} & ICCV-21 & 63.40 & - & 87.80 & - & - & - & - & - & - & - & - & - \\ 
    SRCF~\cite{suo2022simple} & ECCV-22 & 64.04 & 82.99 & 88.81 & - & 57.18 & 75.01 & 81.49 & - & - & - & - & - \\
    IRRA~\cite{jiang2023cross} 
    & CVPR-23 & 73.38 & 89.93 & 93.71 & 66.13 & 63.46 & 80.25 & 85.82 & 38.06 & 60.20 & 81.30 & 88.20 & 47.17 \\
    RaSa~\cite{bai2023rasa}
    & IJCAI-23 & 76.51 & 90.29 & 94.25 & 69.38 & 65.28 & 80.40 & 85.12 & 41.29 & 66.90 & 86.50 & 91.35 & 52.31 \\
    MACF~\cite{sun2024adaptive} & IJCV-24 & 73.33 & 88.57 & 93.02 & - & 62.95 & 79.93 & 85.04 & - & - & - & - & - \\
    TBPS-CLIP~\cite{cao2024empirical}
    & AAAI-24 & 73.54 & 88.19 & 92.35 & 65.38 & 65.05 & 80.34 & 85.47 & 39.83 & 61.95 & 83.55 & 88.75 & 48.26 \\
    AUL$^*$~\cite{li2024adaptive}
    & AAAI-24 & 77.23 & 90.43 & 94.41 & - & 69.16 & 83.32 & 88.37 & - & 71.65 & 87.55 & 92.05 & - \\
    RDE~\cite{qin2024noisy}
    & CVPR-24 & 75.94 & 90.14 & 94.12 & 67.56 & 67.68 & 82.47 & 87.36 & 40.06 & 65.35 & 83.95 & 89.90 & 50.88 \\
    PTMI~\cite{lu2025prompt} & TIFS-25 & 76.02 & 89.93 & 94.14 & 70.85 & 66.54 & 81.50 & 86.50 & 46.05 & 63.35 & 81.95 & 89.10 & 52.46 \\
    GAHR~\cite{qi2025granularity} & TIFS-25 & 76.64 & 90.46 & 94.10 & 66.81 & 68.69 & 82.83 & 87.40 & 42.10 & 68.85 & 86.50 & 91.10 & 53.60 \\
    ICL~\cite{qin2025human} & CVPR-25 & 78.18 & 91.63 & 94.83 & 69.58 & 69.22 & 83.49 & 88.06 & 42.34 & 70.00 & 86.60 & 91.70 & 54.16 \\
    \hline
    \multicolumn{14}{l}{\textbf{\textit{Weakly-supervised TBPS}}} \\
    \hline
    CMMT~\cite{zhao2021weakly}
    &  ICCV-21 & 57.10 & 78.14 & 85.23 & - & - & - & - & - & - & - & - & - \\
    CPCL~\cite{zheng2024cpcl}
    & arXiv-24 & 70.03 & 87.28 & 91.78 & 63.19 & 62.60 & 79.09 & 84.46 & 36.16 & 58.35 & 81.05 & 87.65 & 45.81 \\
    DG-CMIA~\cite{zhang2025dual}
    & arXiv-25 & 73.06 & 89.21 & 93.44 & 64.88 & 63.71 & 79.90 & 85.38 & 36.87 & 61.30 & 82.00 & 88.10 & 47.20 \\
    \hline
    \multicolumn{14}{l}{\textbf{\textit{Semi-supervised TBPS}}} \\
    \hline
    STBPS~\cite{gao2024semi}
    &  arXiv-24 & 63.87 & 82.20 & 87.70 & 57.18 & 46.46 & 64.34 & 71.60 & 26.90 & 56.45 & 78.95 & 87.05 & 44.45 \\
    \hline
    \multicolumn{14}{l}{\textbf{\textit{Unsupervised TBPS}}} \\
    \hline
    GTR~\cite{bai2023text} & ACMMM-23 & 47.53 & 68.23 & 75.91 & 42.91 & 28.25 & 45.21 & 53.51 & 13.82 & 45.60 & 70.35 & 79.95 & 33.30 \\
    GAAP~\cite{li2024cross} & IJCAI-24 & 47.64 & 67.79 & 76.08 & 41.28 & 27.12 & 44.91 & 53.56 & 11.43 & 44.45 & 65.15 &  75.30 & 31.21 \\
    MUMA~\cite{li2025exploring} & AAAI-25 & 59.52 & 77.79 & 84.65 & 52.75 & 38.11 & 56.01 & 63.96 & 19.02 & 54.35 & 76.05 & 83.65 & 40.50 \\
    \rowcolor[HTML]{D7F6FF}
    GTR+ & Ours & 61.35 & 79.35 & 85.75 & 55.75 & 47.81 & 64.97 & 71.94 & 28.75 & 54.75 & 75.15 & 83.50 & 43.79 \\
    \rowcolor[HTML]{D7F6FF}
    GTR+(Pre-training)  & Ours & 64.65 & 80.72 & 86.78 & 58.67 & 52.78 & 67.94 & 73.91 & 33.99 & 55.70 & 76.55 & 84.25 & 43.86 \\
    \hline\thickhline
    \end{tabular}
    }
    \caption{Performance comparison of TBPS methods under different settings. Supervised TBPS uses a large number of images with manually annotated texts for training. Weakly-supervised TBPS only requires image-text pairs, without identity annotations. Semi-supervised TBPS leverages a small set of annotated images ($1\%$ of all training data) and a large set of unannotated ones. Unsupervised TBPS operates without any textual annotations. GTR$^+$(Pre-training) denotes that our proposed GTR$^+$ can be pre-trained on our LargeFine-Person dataset.}
    \label{tab:result_comparision}
\end{table*}

\section{Experiment}
\subsection{Datasets}

Our experiments involve three open benchmark datasets and one proposed pre-training dataset. 

\textbf{CUHK-PEDES}~\cite{li2017person} is the most commonly-used dataset in TBPS. It consists of $40,206$ images and $80,440$ texts from $13,003$ identities in total, which are split into $34,054$ images and $68,126$ texts from $11,003$ identities in the training set, $3,078$ images and $6,158$ texts from $1,000$ identities in the validation set, and $3,074$ images and $6,156$ texts from $1,000$ identities in the test set. The average length of all texts is $23$.

\textbf{ICFG-PEDES}~\cite{ding2021semantically} contains $54,522$ images from $4,102$ identities in total. Each of the images is described by one text. The dataset is split into $34,674$ images from $3,102$ identities in the training set, and $19,848$ images from $1,000$ identities in the test set. On average, there are $37$ words for each text.

\textbf{RSTPReid}~\cite{zhu2021dssl} consists of $20,505$ images of $4,101$ identities. Each identity has $5$ corresponding images captured from different cameras. Each image is annotated with $2$ textual descriptions, and each description is no shorter than $23$ words. There are $3,701$/$200$/$200$ identities utilized for training/validation/testing, respectively.

\textbf{LargeFine-Person} is our proposed large-scale, fine-grained description dataset for pre-training TBPS. It comprises $1.2$M images covering $611,460$ identities, with each image annotated with $3$ textual descriptions on average. The descriptions exhibit rich linguistic variation and fine-grained detail, with an average length of $61$ words per description, and up to $507$ words in total.

During the experiments, unless otherwise specified, we evaluate the proposed GTR+ on each dataset under an unsupervised setting. Specifically, only the images from the training set are used for training, while the corresponding textual annotations in the training set are excluded during the training process.

\subsection{Protocol}
We adopt the widely-used Rank@K (R@K for short, K=1, 5, 10) metric to evaluate the performance of the proposed method. Specifically, given a query text, we rank all the test images via the similarity with the query text, and the search is deemed to be successful if top-K images contain any corresponding identity. R@K is the percentage of successful searches. In addition, we also adopt the mean average precision (mAP) as a complementary metric. Rank@K reflects the accuracy of the first few retrieval results, while mAP emphasizes the comprehensive performance of the method.

\subsection{Implementation Details}
We train the retrieval model using image inputs resized to $256 \times 256$ and textual inputs truncated to a maximum of $77$ tokens.  
The model is optimized using the Adam optimizer with a learning rate of $1e\text{-}6$, a batch size of $25$, and trained for $15$ epochs.  
All experiments are conducted on NVIDIA L40 GPUs.  
In addition, the hyper-parameter $\gamma$ in Eq.~\ref{eq-sum} is set to $0.2$, and $\beta$ in Eq.~\ref{eq:loss_itc} and Eq.~\ref{eq:loss_itm} is set to $0.8$.

\begin{table*}[t]
    \centering
    \resizebox{1\linewidth}{!}{
    \renewcommand\arraystretch{1.2}
    \begin{tabular}{c|l|c||ccccIccccIcccc}
    \hline\thickhline
    \rowcolor{lightgray}
    & & & \multicolumn{4}{cI}{CUHK-PEDES} & \multicolumn{4}{cI}{ICFG-PEDES} & \multicolumn{4}{c}{RSTPReid}  \\
    \rowcolor{lightgray}
    \multirow{-2}{*}{Pre-training Datasets}
    & \multirow{-2}{*}{Baselines} & \multirow{-2}{*}{Fine-tuning} & R@1 & R@5 & R@10 & mAP & R@1 & R@5 & R@10 & mAP & R@1 & R@5 & R@10 & mAP \\
    \hline\hline

    &   & \texttimes 
    & 20.32 & 38.09 & 48.38 & 19.14
    & 9.10 & 20.64 & 28.24 & 3.86
    & 22.20 & 44.55 & 58.15 & 17.03 \\
    & \multirow{-2}{*}{RDE~\cite{qin2024noisy}} 
    & \checkmark 
    & 53.51 & 73.80 & 81.69 & 48.23
    & 42.44 & 58.98 & 66.33 & 21.88
    & 51.80 & 72.80 & 80.95 & 38.75\\
    \cline{2-15} 
    &   & \texttimes    
    & 21.10 & 38.54 & 48.52 & 19.51 
    & 10.67 & 21.95 & 28.19 & 4.73 
    & 22.51 & 45.68 & 59.36 & 17.11 \\   
    & \multirow{-2}{*}{IRRA~\cite{jiang2023cross}} 
    & \checkmark 
    & 57.65 & 75.81 & 82.34 & 51.46 
    & 41.12 & 59.13 & 67.07 & 21.96 
    & 49.10 & 72.85 & 81.45 & 37.52 \\
    \cline{2-15}
    &   & \texttimes 
    & 26.98 & 46.18 & 55.36 & 23.59 
    & 12.47 & 22.04 & 28.22 & 5.30 
    & 26.15 & 52.20 & 64.20 & 19.15 \\
    \multirow{-6}{*}{LUPerson-T~\cite{shao2023unified}} 
    & \multirow{-2}{*}{BLIP~\cite{li2022blip}} 
    & \checkmark 
    & 61.16 & 79.55 & 86.35 & 55.75 
    & 45.84 & 63.30 & 70.15 & 26.65 
    & 51.30 & 74.50 & 81.85 & 41.02 \\
    \hline

    &   & \texttimes 
    & 54.39 & 73.02 & 80.34 & 48.48 
    & 36.41 & 54.99 & 63.49 & 18.84
    & 49.80 & 72.45 & 82.00 & 37.05 \\
    & \multirow{-2}{*}{RDE~\cite{qin2024noisy}} 
    & \checkmark 
    & 55.88 & 76.32 & 82.94 & 49.73
    & 42.80 & 60.20 & 67.61 & 22.88
    & 52.80 & 73.80 & 83.00 & 39.29 \\
    \cline{2-15} 
    &   & \texttimes 
    & 57.60 & 75.94 & 82.77 & 51.44 
    & 38.30 & 56.60 & 64.54 & 20.43 
    & 51.50 & 73.92 & 82.65 & 37.34 \\
    & \multirow{-2}{*}{IRRA~\cite{jiang2023cross}} 
    & \checkmark 
    & 60.41 & 79.56 & 85.51 & 54.24 
    & 46.00 & 62.45 & 69.63 & 24.67 
    & 54.70 & 75.40 & 84.10 & 40.14 \\
    \cline{2-15} 
    &   & \texttimes 
    & 60.11 & 77.45 & 84.15 & 53.46 
    & 41.97 & 59.75 & 67.01 & 20.81 
    & 52.40 & 75.15 & 82.80 & 38.81 \\
    \multirow{-6}{*}{LUPerson-MLLM~\cite{tan2024harnessing}}
    & \multirow{-2}{*}{BLIP~\cite{li2022blip}} 
    & \checkmark 
    & \underline{64.52} & \underline{80.07} & \textbf{86.91} & \underline{57.74} 
    & \underline{49.89} & \underline{66.40} & \underline{73.12} & \underline{30.73} 
    & \underline{54.90} & \underline{76.20} & \underline{83.85} & \textbf{44.67} \\
    \hline

    \rowcolor[HTML]{D7F6FF}
    &   & \texttimes 
    & 56.45 & 75.05 & 82.10 & 50.05 
    & 41.05 & 59.15 & 66.75 & 20.78 
    & 50.70 & 73.65 & 82.40 & 37.15 \\
    \rowcolor[HTML]{D7F6FF}
    & \multirow{-2}{*}{RDE~\cite{qin2024noisy}} 
    & \checkmark 
    & 56.97 & 75.93 & 82.78 & 50.78
    & 44.47 & 60.71 & 67.75 & 23.78
    & 50.85 & 72.15 & 81.70 & 38.42 \\
    \hhline{|~|--------------|}
    \rowcolor[HTML]{D7F6FF}
    &   & \texttimes 
    & 59.44 & 78.54 & 85.22 & 54.11
    & 43.77 & 60.77 & 68.05 & 22.30 
    & 50.45 & 73.45 & 82.35 & 37.68 \\
    \rowcolor[HTML]{D7F6FF}
    & \multirow{-2}{*}{IRRA~\cite{jiang2023cross}} 
    & \checkmark 
    & 60.88 & 80.02 & 86.45 & 55.54
    & 46.02 & 62.76 & 69.92 & 24.58   
    & 54.40 & 76.05 & 84.10 & 41.31 \\
    \hhline{|~|--------------|}
    \rowcolor[HTML]{D7F6FF}
    &   & \texttimes 
    & 62.65 & 78.80 & 84.76 & 55.27 
    & 47.53 & 64.32 & 71.39 & 25.38 
    & 52.00 & 74.05 & 82.35 & 38.72 \\
    \rowcolor[HTML]{D7F6FF}
    \multirow{-6}{*}{LargeFine-Person (Ours)} 
    & \multirow{-2}{*}{BLIP~\cite{li2022blip}} 
    & \checkmark 
    & \textbf{64.65} & \textbf{80.72} & \underline{86.78} & \textbf{58.67} 
    & \textbf{52.78} & \textbf{67.94} & \textbf{73.91} & \textbf{33.99} 
    & \textbf{55.70} & \textbf{76.55} & \textbf{84.25} & \underline{43.86} \\
    \hline\thickhline
    \end{tabular}
    }
    \caption{Comparison of pre-training datasets under the unsupervised setting. 
    Only two pre-training datasets, LUPerson-T and LUPerson-MLLM, conform to the unsupervised setting (see Table~\ref{tab:data_comparision}) and are used for a fair comparison.
    $\times$Fine-tuning: the baseline is only trained on the pre-training dataset and evaluated directly. $\checkmark$Fine-tuning: the baseline is further fine-tuned on the respective downstream dataset (images from the original dataset paired with synthesized descriptions via our tiered description generation framework). Best and second-best results are denoted in \textbf{bold} and \underline{underline}, respectively.
}
    \label{tab:pretraining_comparision_unsup}
    \vspace{-0.25cm}
\end{table*}

\subsection{Comparisons with the State-of-the-Art Methods}

We compare the proposed GTR+ with SOTA methods on CUHK-PEDES, ICFG-PEDES and RSTPReID in Table~\ref{tab:result_comparision}. 
These SOTA methods fall into four distinct learning paradigms.
Supervised TBPS relies on fully annotated image-text pairs; weakly supervised TBPS requires image-text pairs without identity labels; semi-supervised TBPS leverages a small annotated set (\emph{i.e.}, $1\%$ of data) and a large amount of unannotated images; unsupervised TBPS operates entirely without textual annotations.
Among these settings, the unsupervised paradigm provides the most fair and direct comparison with our GTR+. In contrast, the other paradigms leverage additional supervision signals, which naturally confer a performance advantage.
As expected, GTR+ exhibits a performance gap compared to methods that utilize more supervision signals, due to their access to richer annotated data. 
Whereas, compared to unsupervised methods, GTR+ achieves significantly superior performance, demonstrating its effectiveness.
Notably, when further pre-trained on our constructed LargeFine-Person dataset—still under an unsupervised setting—GTR+ achieves even stronger results, even outperforming semi-supervised method. This highlights the potential of our constructed large-scale dataset.

\begin{table}[t]
    \centering
    \resizebox{0.9\linewidth}{!}{
    \renewcommand\arraystretch{1.2}
    \begin{tabular}{c|c||cccc}
    \hline\thickhline
    \rowcolor{lightgray}
    & Pre-training Datasets & R@1 & R@5 & R@10 & mAP \\
    \hline\hline
    \multirow{8}{*}{\rotatebox{90}{CUHK-PEDES}}
    & \multicolumn{5}{c}{w/ Human-annotation} \\
    \cline{2-6}
    & MALS~\cite{yang2023towards} 
    & 74.05 & 89.48 & 93.64 & 66.57  \\
    & SYNTH-PEDES~\cite{zuo2024plip} 
    & 74.88 & 89.58 & 94.29 & 67.10 \\
    & HAM-PEDES~\cite{jiang2025modeling} 
    & 77.71 & 91.42 & 94.57 & 69.68 \\
    \cline{2-6}
    & \multicolumn{5}{c}{w/o Human-annotation} \\
    \cline{2-6}
    & LUPerson-T~\cite{shao2023unified} 
    & 74.37 & 89.51 & 93.97 & 66.60  \\
    & LUPerson-MLLM~\cite{tan2024harnessing} 
    & {76.82} & {91.16} & {94.46} & {69.55}  \\
    \rowcolor[HTML]{D7F6FF}
    & LargeFine-Person (Ours) 
    & {77.13} & {90.82} & {94.49} & {68.37} \\
    \hline\hline
    \multirow{8}{*}{\rotatebox{90}{ICFG-PEDES}}
    & \multicolumn{5}{c}{w/ Human-annotation} \\
    \cline{2-6}
    & MALS~\cite{yang2023towards} 
    & 64.37 & 80.75 & 86.12 & 38.85  \\
    & SYNTH-PEDES~\cite{zuo2024plip}          
    & 64.31 & 80.92 & 86.53 & 38.49 \\
    & HAM-PEDES~\cite{jiang2025modeling} 
    & 68.25 & 83.30 & 88.15 & 42.30 \\
    \cline{2-6}
    & \multicolumn{5}{c}{w/o Human-annotation} \\
    \cline{2-6}
    & LUPerson-T~\cite{shao2023unified} 
    & 64.50 & 80.24 & 85.74 & 38.22  \\
    & LUPerson-MLLM~\cite{tan2024harnessing} 
    & {67.05} & {82.16} & {87.33} & {41.51} \\
    \rowcolor[HTML]{D7F6FF}
    & LargeFine-Person (Ours)                
    & {67.80} & {82.81} & {87.66} & {41.00} \\
    \hline\hline
    \multirow{8}{*}{\rotatebox{90}{RSTPReid}}
    & \multicolumn{5}{c}{w/ Human-annotation} \\
    \cline{2-6}
    & MALS~\cite{yang2023towards}             
    & 61.90 & 80.60 & 89.30 & 48.08 \\
    & SYNTH-PEDES~\cite{zuo2024plip}      
    & 64.00 & 82.85 & 89.90 & 50.06 \\
    & HAM-PEDES~\cite{jiang2025modeling}  
    & 71.69 & 87.85 & 93.30 & 55.19 \\
    \cline{2-6}
    & \multicolumn{5}{c}{w/o Human-annotation} \\
    \cline{2-6}
    & LUPerson-T~\cite{shao2023unified}      
    & 62.20 & 83.30 & 89.75 & 48.33 \\
    & LUPerson-MLLM~\cite{tan2024harnessing}  
    & {68.50} & {87.15} & {92.10} & {53.02} \\
    \rowcolor[HTML]{D7F6FF}
    & LargeFine-Person (Ours)            
    & {69.05} & {86.90} & {92.25} & {54.19} \\
    \hline\thickhline
    \end{tabular}
    }
    \caption{Comparison of pre-training datasets under supervised setting. We adopt IRRA~\cite{jiang2023cross} as the retrieval baseline.}
    \label{tab:pretraining_comparision_sup}
    \vspace{-0.5cm}
\end{table}

Furthermore, we compare the proposed LargeFine-Person pre-training dataset with other pre-training datasets in terms of performance, 
To this end, we pre-train various retrieval baselines on each dataset, and the results are presented in Table~\ref{tab:pretraining_comparision_unsup} and Table~\ref{tab:pretraining_comparision_sup}.
\ding{182} We first conduct comparisons under unsupervised setting. Notably, only LUPerson-T and LUPerson-MLLM are constructed independently of human annotations, thus aligning with the unsupervised paradigm. For a fair comparison, we compare our dataset against these two pre-training datasets.
As shown in Table~\ref{tab:pretraining_comparision_unsup}, when the retrieval baseline is trained on the pre-training dataset and evaluated directly (i.e., without fine-tuning), our LargeFine-Person consistently achieves superior performance across all baselines, demonstrating its effectiveness.
Furthermore, owing to our proposed tiered description generation framework, downstream datasets—CUHK-PEDES, ICFG-PEDES, and RSTPReid—can be expanded by synthesizing paired textual descriptions using only image data, fully adhering to the unsupervised setting. The baseline models are then fine-tuned on these synthesized datasets (i.e., with fine-tuning). It can be seen that LargeFine-Person continues to outperform the competing datasets across all baselines in this scenario as well.
Moreover, we observe a significant performance gain across all baselines after fine-tuning, which further validates the high-quality of the synthesized downstream dataset and the effectiveness of our tiered description generation framework.
\ding{183} We further conduct comparisons with other pre-training datasets under the supervised setting. In this setup, the retrieval model is first pre-trained on the pre-training dataset and then conventionally fine-tuned on the downstream dataset.
As shown in Table~\ref{tab:pretraining_comparision_sup}, the model pre-trained on our LargeFine-Person achieves competitive performance compared to other unsupervised pre-training datasets (i.e., those constructed without human annotation). However, it is slightly inferior to pre-training datasets that leverage human-annotated data during construction—particularly on the HAM-PEDES dataset.
This performance gap is reasonable: HAM-PEDES is generated with reference to human-annotated textual descriptions, which ensures higher semantic fidelity. In contrast, LargeFine-Person is constructed entirely without human supervision. As a result, HAM-PEDES contains higher-quality image-text pairs, leading to stronger pre-training efficacy.

\subsection{Ablation Study}

\begin{table*}[t]
    \centering
    \resizebox{0.78\linewidth}{!}{
        \renewcommand\arraystretch{1.2}

    \begin{tabular}{lIccIccIcc||cccc}
    \hline\thickhline
    \rowcolor{lightgray}
    & \multicolumn{2}{cI}{Base tier} & \multicolumn{2}{cI}{Intermediate tier} & \multicolumn{2}{c||}{Advanced tier} &  &  &  &  \\
    \rowcolor{lightgray}
    \multirow{-2}{*}{Methods} & Temp. & \textbf{Q\&A} & Rand. & \textbf{ISC} & Intg. & \textbf{StyExp}  & \multirow{-2}{*}{R@1} & \multirow{-2}{*}{R@5} & \multirow{-2}{*}{R@10} & \multirow{-2}{*}{mAP} \\
    \hline\hline
    TierG\_Base$_T$
    &  \checkmark &  &  &  &  &  & 47.27 & 67.58 & 75.62 & 42.28 \\
    TierG\_Base$_Q$
    &   & \checkmark &  &  &  &  & 54.82 & 73.91 & 80.99 & 49.42 \\
    \hline
    TierG\_Interm$_R$
    &   &  & \checkmark &  &  &  & 53.33 & 73.16 & 80.49 & 47.80 \\
    TierG\_Interm$_I$
    &   &  &  & \checkmark &  &  & 54.58 & 73.85 & 81.03 & 49.31 \\
    \hline
    TierG$_I$
    &   &  \checkmark &  & \checkmark & \checkmark &  & 55.23 & 74.24 & 81.51 & 49.72 \\
    \rowcolor[HTML]{D7F6FF}
    TierG
    &   &  \checkmark &  & \checkmark &  & \checkmark & 55.58 & 74.40 & 81.87 & 50.12 \\
    \hline\thickhline
    \end{tabular}
    }
    \caption{Ablation Study of the proposed tiered description generation framework (TierG) on CUHK-PEDES. Q\&A: question-and-answer mechanism; ISC: inter-sample contrastive mechanism; StyExp: stylized expansion mechanism. Temp.: template-based generation (vs. Q\&A); Rand.: random negative sampling (vs. ISC); Intg.: direct integration without stylized expansion (vs. StyExp).}
    \label{tab:ablation_three}
\end{table*}

\begin{table}[t]
    \centering
    \resizebox{0.76\linewidth}{!}{
        \renewcommand\arraystretch{1.2}
    \begin{tabular}{l||cccc}
    \hline\thickhline
    \rowcolor{lightgray}
    & R@1 & R@5 & R@10 & mAP \\
    \hline\hline
    Baseline
    & 55.58 & 74.40 & 81.87 & 50.12\\
    \hline
    k=1
    & 57.94 & 76.92 & 83.77 & 51.66\\
    k=5
    & 58.45 & 77.25 & 84.01 & 52.01 \\
    \rowcolor[HTML]{D7F6FF}
    k=10 (Ours)
    & \textbf{59.50} & \textbf{78.49} & \textbf{84.67} & \textbf{53.56} \\
    \hline\thickhline
    \end{tabular}
    }
    \caption{Sensitivity analysis of the top-$k$ choice for hard negative selection in the intermediate tier on CUHK-PEDES. Best results are denoted in bold.}
    \label{tab:ablation-topk}
\end{table}

\subsubsection{Effectiveness of the tiered description generation framework}

We propose a tiered description generation framework (TierG) for textual description generation, which incorporates three key components: the base tier with the question-and-answer mechanism (Q\&A), the intermediate tier with the inter-sample contrastive mechanism (ISC), and the advanced tier with the stylized expansion mechanism (StyExp).
In this part, we conduct an ablation study to evaluate the contribution of each component, as summarized in Table~\ref{tab:ablation_three}.
\ding{182} The goal of the base tier is to generate a basic textual description.
In previous work~\cite{bai2023text}, we employ a template-based approach (Temp.), in which the instruction is used to prompt the vision-language model to extract attributes, followed by filling these attributes into a manually designed template to construct the final description.
This serves as a comparison for our proposed Q\&A mechanism.
Compared to Temp., Q\&A offers clear performance advantages.
Since Q\&A provides greater flexibility in attribute generation and eliminates the need for laborious manual template design, enabling more natural and varied descriptions.
\ding{183} For the intermediate tier, the objective is to generate fine-grained textual descriptions.
For this, we introduce hard negative images selected from the top-$k$ most similar images as comparison during the generation process (ISC).
In contrast, randomly selecting a negative image from all images (Rand.) is a feasible solution, but leads to the degradation in performance.
The use of hard negatives in ISC provides more informative contrastive signals, which significantly benefits the model in generating more precise and discriminative descriptions.
\ding{184} In the advanced tier, we aim to generate descriptions with diversified styles. For this, we design an instruction that builds upon the outputs from the first two tiers, guided by an explicitly stylized template to steer the MLLM toward generating stylistically rich text (StyExp).
Alternatively, we can omit the stylized template and simply integrate the descriptions from the base and intermediate tiers for generation (Intg.).
Results show that StyExp achieves superior performance, demonstrating that explicitly diversifying the style of textual descriptions is effective for improving TBPS performance.
\ding{185} Overall, our proposed TierG employs a three-tier sequential architecture to generate high-quality textual descriptions in a progressive manner. 
Results show that TierG outperforms TierG\_Base$_Q$ and TierG\_Interm$_I$, which strongly validates the effectiveness of the tiered design. 

\begin{table*}[t]
    \centering
    \resizebox{0.88\linewidth}{!}{
        \renewcommand\arraystretch{1.2}
        
    \begin{tabular}{lIcccIcc||cccc}
    \hline\thickhline
    \rowcolor{lightgray}
    & \multicolumn{3}{cI}{Masking} & \multicolumn{2}{c||}{Weighting} &  &  &  &  \\
    \rowcolor{lightgray}
    \multirow{-2}{*}{Methods} & Text (sim.) & Text (prob.) & Image & Fixed conf. & Adaptive conf. & \multirow{-2}{*}{R@1} & \multirow{-2}{*}{R@5} & \multirow{-2}{*}{R@10} & \multirow{-2}{*}{mAP} \\
    \hline\hline
    Baseline & & & & & & 55.58 & 74.40 & 81.87 & 50.12 \\
    \hline
    Sim-TxtMask~\cite{tan2024harnessing} & \cmark & & & & & 55.60 & 74.45 & 81.93 & 49.26 \\
    Prob-TxtMask &  & \cmark & & & & 55.67 & 74.51 & 82.13 & 50.76 \\
    Rand-ImgMask &  & & \cmark & & & 56.68 & 75.62 & 82.34 & 51.26 \\
    \hline    
    FixWeight~\cite{bai2023text} &  & & & \cmark & & 58.72 & 77.71 & 84.28 & 54.10 \\
    \rowcolor[HTML]{D7F6FF}
    AdaWeight&  & & &  & \cmark & 61.35 & 79.35 & 85.75 & 55.75  \\
    \hline\thickhline
    \end{tabular}
    }
    \caption{Ablation Study of the proposed adaptive confidence-weighted retrieval learning framework on CUHK-PEDES. Noise-robust learning can be achieved through two main strategies. (1) Masking noisy tokens/patches, including: masking text tokens with low image-text similarity scores (Text (sim.)), masking text tokens with low generation probabilities (Text (prob.)), randomly masking image patches (Image); (2) Weighting training samples based on noise levels, including: Fixed confidence weighting (Fixed conf.), Adaptive confidence weighting (Adaptive conf.).}
    \label{tab:ablation_noise}
\end{table*}

\begin{table}[t]
    \centering
    \resizebox{0.76\linewidth}{!}{
        \renewcommand\arraystretch{1.2}
    \begin{tabular}{l||cccc}
    \hline\thickhline
    \rowcolor{lightgray}
    & R@1 & R@5 & R@10 & mAP \\
    \hline\hline
    Baseline
    & 55.58 & 74.40 & 81.87 & 50.12 \\
    \hline
    $K$=1
    & 57.29 & 77.23 & 83.56 & 51.21 \\
    \rowcolor[HTML]{D7F6FF}
    $K$=2 (Ours)
    & \textbf{61.35} & \textbf{79.35} & \textbf{85.75} & \textbf{55.75} \\
    $K$=3
    & 59.31 & 78.46 & 84.84 & 53.54 \\    
    $K$=4
    & 59.49 & 78.40 & 84.68 & 53.79 \\    
    $K$=5
    & 59.34 & 78.62 & 84.78 & 53.61 \\
    \hline\thickhline
    \end{tabular}
    }
    \caption{Sensitivity analysis of the number of Gaussian components in GMM-based confidence estimation on CUHK-PEDES. $K$ denotes the number of Gaussian components. Best results are denoted in bold.}
    \label{tab:ablation-gmm-k}
    \vspace{-0.25cm}
\end{table}

\subsubsection{Effectiveness of the adaptive confidence-weighted retrieval learning framework}
We propose an adaptive confidence-weighted retrieval learning framework (AdaWeight), to mitigate the impact of noisy data introduced during the text generation stage. To evaluate the effectiveness of AdaWeight, we compare with several alternative strategies.
Tan et al.~\cite{tan2024harnessing} introduced LUPerson-MLLM, a large-scale TBPS dataset constructed using synthetic textual descriptions, and proposed a noise-robust retrieval method by masking text tokens with low image-text similarity scores. 
Inspired by this work, we observe that besides the weighting strategy, the masking strategy can also serve as an effective means for noise-robust learning.
We present the comparison results in Table~\ref{tab:ablation_noise}.
\ding{182} Masking strategy for noise-robust learning. We evaluate three masking approaches:
(i) Following~\cite{tan2024harnessing}, we mask text tokens with low similarity scores relative to the paired images (Sim-TxtMask);
(ii) Alternatively, we mask tokens with low generation probabilities from the generation model (Prob-TxtMask);
(iii) We extend this idea to the image domain by randomly masking image patches (Rand-ImgMask), which can implicitly reduce the influence of noisy correspondences.
From Table~\ref{tab:ablation_noise}, Rand-ImgMask achieves the best performance, showing a slight improvement over text-based masking variants.
The image typically exhibits more semantic coherence than the text and retains more meaningful information after partial masking, enabling more robust cross-modal alignment under noisy conditions.
\ding{183} Weighting strategy for noise-robust learning. In previous work~\cite{bai2023text}, a fixed confidence-weighting approach (FixWeight) is proposed based on text generation probabilities: higher probabilities indicate more confident generations, implying less noise, and thus greater weights. However, FixWeight fails to capture dynamic changes during training, potentially leading to suboptimal handling of noisy samples. In contrast, our proposed adaptive confidence-weighted approach (AdaWeight) dynamically adjusts sample importance according to both generation confidence and training progress, offering a more principled and flexible solution. 
\ding{184} Masking vs. Weighting. Overall, the weighting strategy demonstrates superior effectiveness compared to the masking strategy, with our proposed AdaWeight achieving the best performance and offering clear advantages.

\begin{figure}[!b]
  \centering
\includegraphics[width=\linewidth]{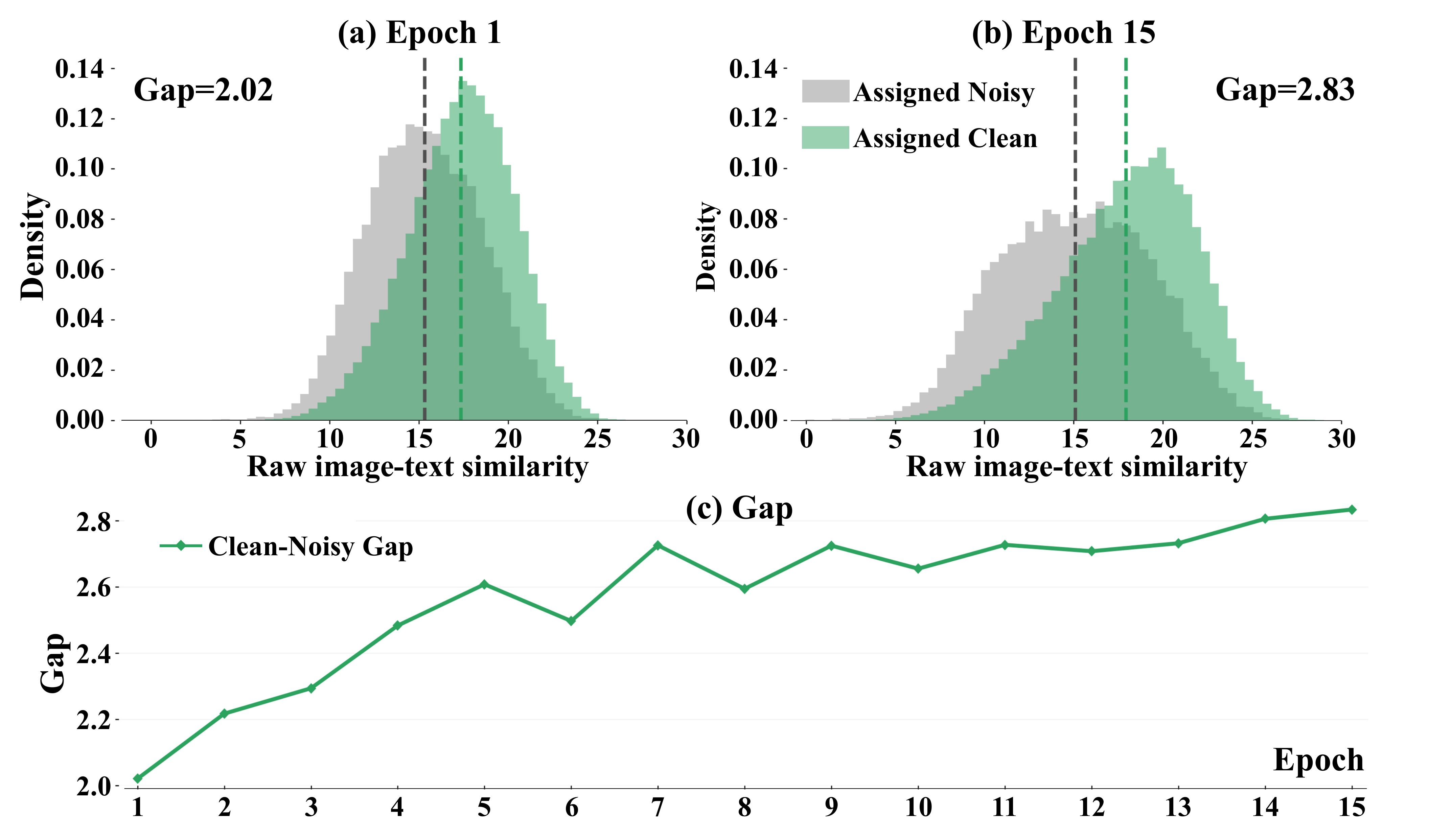}
    \caption{Visualization of GMM-based separation during training. Panels (a) and (b) show the distributions of image-text similarity for the subsets assigned by the GMM to the noisy and clean components at epochs 1 and 15, respectively. Panel (c) shows the gap between the clean and noisy subsets over training epochs.}
  \label{fig:gmm analysis}
 \end{figure}

\begin{figure}[htbp]
  \centering
\includegraphics[width=\linewidth]{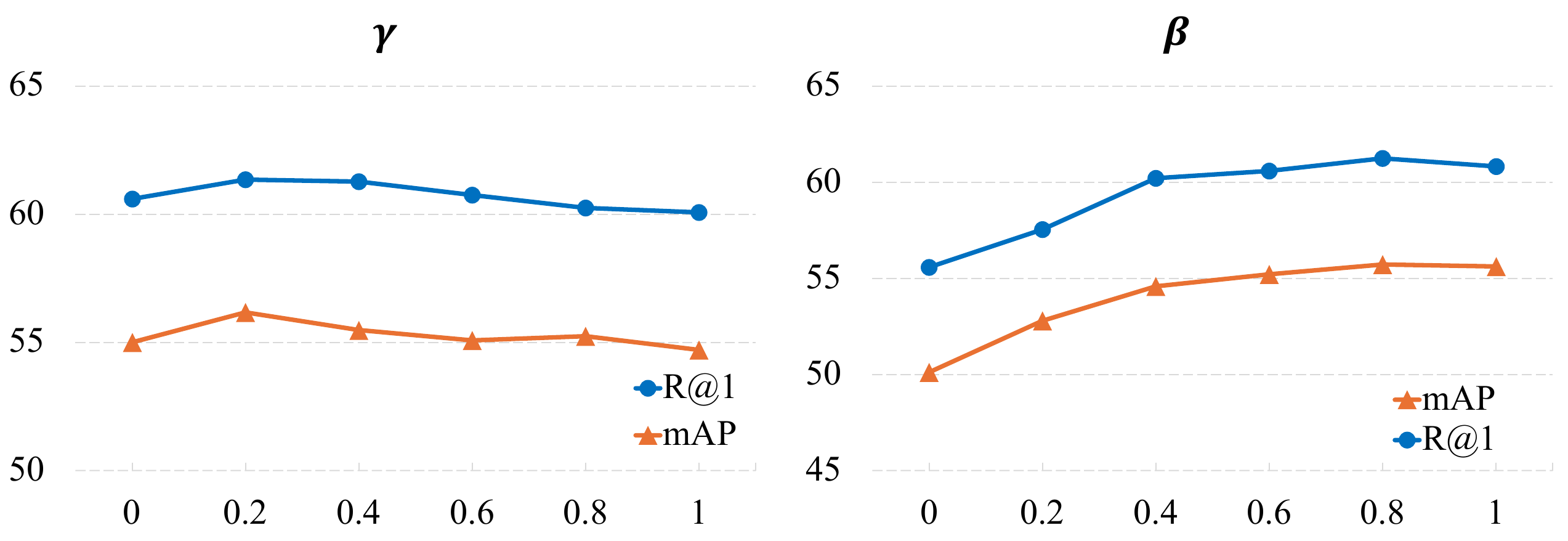}
    \caption{Hyper-parameter analysis of $\gamma$ and $\beta$.}
  \label{fig:Hyper-parameter analysis}
\end{figure}

\subsubsection{Hyper-parameter analysis}

\ding{182} 
\textbf{Top-$k$ hard negative selection.} We evaluate the sensitivity of the top-$k$ hard negative selection during the intermediate generation tier, as shown in Table.~\ref{tab:ablation-topk}. While all settings outperform the baseline, $k=10$ yields the best results. An overly small $k$ (e.g., $k=1$) restricts candidates to a narrow neighborhood, limiting comparison diversity. Conversely, sampling from the top-10 neighbors strikes an optimal balance: it preserves visual similarity while providing diverse contrastive cues, which generates richer discriminative descriptions and improves retrieval performance.
\ding{183} 
\textbf{Gaussian components ($K$) and GMM discriminative capability.} To explicitly validate the discriminative capability of our GMM, Figure.~\ref{fig:gmm analysis} visualizes its stage-wise behavior. The model consistently separates training pairs into clean and noisy subsets. More importantly, the similarity gap between these two subsets gradually widens as training proceeds. This dynamic confirms that the GMM is not merely fitting a static bimodal shape, but progressively and reliably identifying clean samples as the feature representations improve. Furthermore, we justify our two-component design by analyzing the sensitivity of the Gaussian component number ($K$), as reported in Table.~\ref{tab:ablation-gmm-k}. The results demonstrate that $K=2$ yields the optimal performance. A single component ($K=1$) is too coarse to model the underlying noise variation, whereas using more components ($K>2$) tends to over-partition the distribution, leading to degraded confidence estimation.
\ding{184} 
\textbf{Impact of $\gamma$ and $\beta$.} Our adaptive confidence-weighted retrieval learning framework involves two hyper-parameters: $\gamma$ and $\beta$.  
The parameter $\gamma$ balances the contributions of pair-wise cosine similarity and text generation probability in computing the confidence scores, while $\beta$ controls the influence of these confidence scores within the training objectives.
A higher $\gamma$ places more weight on cosine similarity, whereas a lower $\gamma$ emphasizes text generation probability. And $\gamma = 1$ corresponds to adopting only cosine similarity for computing the confidence scores, and $\gamma = 0$ corresponds to using only text generation probability. As shown in Figure~\ref{fig:Hyper-parameter analysis}, relying solely on either component ($\gamma = 0$ or $\gamma = 1$) leads to suboptimal performance. The best results are achieved by combining both with $\gamma = 0.2$. The cosine similarity is an optimizable dynamic value that improves during training and a smaller $\gamma$ helps maintain training stability.
The parameter $\beta$ governs the overall impact of confidence scores on training. When $\beta = 0$, confidence scores have no effect. As illustrated in Figure~\ref{fig:Hyper-parameter analysis}, performance is relatively low when $\beta < 0.5$, becomes more stable for $\beta > 0.5$, and reaches its peak at $\beta = 0.8$.

        


\subsection{Generalization Experiment}

\subsubsection{Domain generalization}
We evaluate the domain generalization capability of the proposed adaptive confidence-weighted learning framework in Table~\ref{tab:cuhk-icfg}.
Specifically, models are trained on the source dataset and directly tested on the target dataset, \emph{e.g.,} C$\rightarrow$I, where training is conducted on CUHK-PEDES and evaluation is performed on ICFG-PEDES.
Notably, all compared methods are supervised approaches that rely on both images and human-annotated textual descriptions during training.
Moreover, among these, MRA~\cite{yang2025minimizing} is TBPS method specifically designed for the domain generalization setting.
In contrast, our GTR+ operates in an unsupervised manner, requiring only images from the source domain, with textual descriptions automatically generated by our proposed tiered description generation framework.
Despite the absence of annotated texts, GTR+ achieves competitive performance, demonstrating the high quality of the synthesized descriptions and the effectiveness of our proposed adaptive confidence-weighted retrieval learning framework.

\subsubsection{Different baseline retrieval models}
\label{sec:Dif-baseline}
We evaluate the compatibility of the proposed adaptive confidence-weighted learning framework by integrating various baseline retrieval models. As shown in Table~\ref{tab:difbaseline}, the framework demonstrates strong adaptability and robustness, consistently achieving competitive performance across different retrieval models.

\begin{table}[t]
\centering
\resizebox{0.78\linewidth}{!}{
    \renewcommand\arraystretch{1.2}
\begin{tabular}{c|l||cccc}
\hline\thickhline\rowcolor{lightgray}
    \multicolumn{2}{c||}{Methods}     & R@1  & R@5  & R@10 & mAP \\
\multirow{6}{*}{\rotatebox[origin=c]{90}{C$\to$I}} 
    & LGUR~\cite{shao2022learning} 
    & 34.25 & 52.28 & 60.85 & - \\
    & IRRA~\cite{jiang2023cross}           
    & 42.62 & 62.39 & 70.13 & 23.89\\
    & BLIP~\cite{li2022blip}        
    & 47.93 & 63.98 & 71.07 & 26.25 \\
    & RDE~\cite{qin2024noisy}
    & 48.15 & 66.36 & 73.43 & 27.01 \\
    & GAHR~\cite{qi2025granularity}
    & 49.00 & 66.34 & 73.59 & - \\
    & MRA*~\cite{yang2025minimizing}  
    & 50.01 & 67.13 & 74.24 & - \\
    \rowcolor[HTML]{D7F6FF}
    & GTR+ (Ours)
    & 48.93 & 65.98 & 71.27 & 28.25\\
\hline\hline
\multirow{6}{*}{\rotatebox[origin=c]{90}{I$\to$C}} 
    & LGUR~\cite{shao2022learning}   
    & 25.44 & 44.48 & 54.39 & - \\
    & IRRA~\cite{jiang2023cross}     
    & 32.21 & 55.12 & 65.19 & 31.72 \\
    & BLIP~\cite{li2022blip}
    & 36.35 & 57.98 & 67.21 & 34.19 \\
    & RDE~\cite{qin2024noisy}
    & 38.33  & 59.51 & 68.30 & 35.56 \\
    & GAHR~\cite{qi2025granularity}
    & 40.60 & 62.46 & 71.62 & - \\ 
    & MRA*~\cite{yang2025minimizing}  
    & 47.66 & 68.70 & 76.82 & - \\
    \rowcolor[HTML]{D7F6FF}
    & GTR+ (Ours)
    & 38.73 & 60.12 & 69.87 & 35.90 \\
\hline\thickhline
\end{tabular}
}
\caption{Domain generalization comparisons. C: CUHK-PEDES, I: ICFG-PEDES. GTR+ trains in source with an unsupervised setting with generated texts, others rely on fully annotated supervised data. * denotes the method specifically designed for the domain generalization scenario.}
\label{tab:cuhk-icfg}
\end{table}

\begin{table}[t]
\centering
\resizebox{0.76\linewidth}{!}{
        \renewcommand\arraystretch{1.2}
\begin{tabular}{l||c|c|c|c}
\hline\thickhline\rowcolor{lightgray}
    Baselines & R@1 & R@5 & R@10 & mAP \\
\hline\hline
    
    RDE~\cite{qin2024noisy}   & 59.93 & 78.89 & 84.18 & 53.69 \\
    IRRA~\cite{jiang2023cross}                & 60.72 & 79.11 & 84.93 & 54.80 \\
    \rowcolor[HTML]{D7F6FF}
    BLIP~\cite{li2022blip}                & 61.35 & 79.35 & 85.75 & 55.75  \\
    
\hline\thickhline
\end{tabular}
}
\caption{Results of adopting different retrieval baselines in the adaptive confidence-weighted learning framework on CUHK-PEDES.}
\label{tab:difbaseline}
\end{table}

\subsection{Visualization Analysis}
\begin{figure*}[htbp]
  \centering
    \includegraphics[width=0.97\linewidth, height=0.32\linewidth]{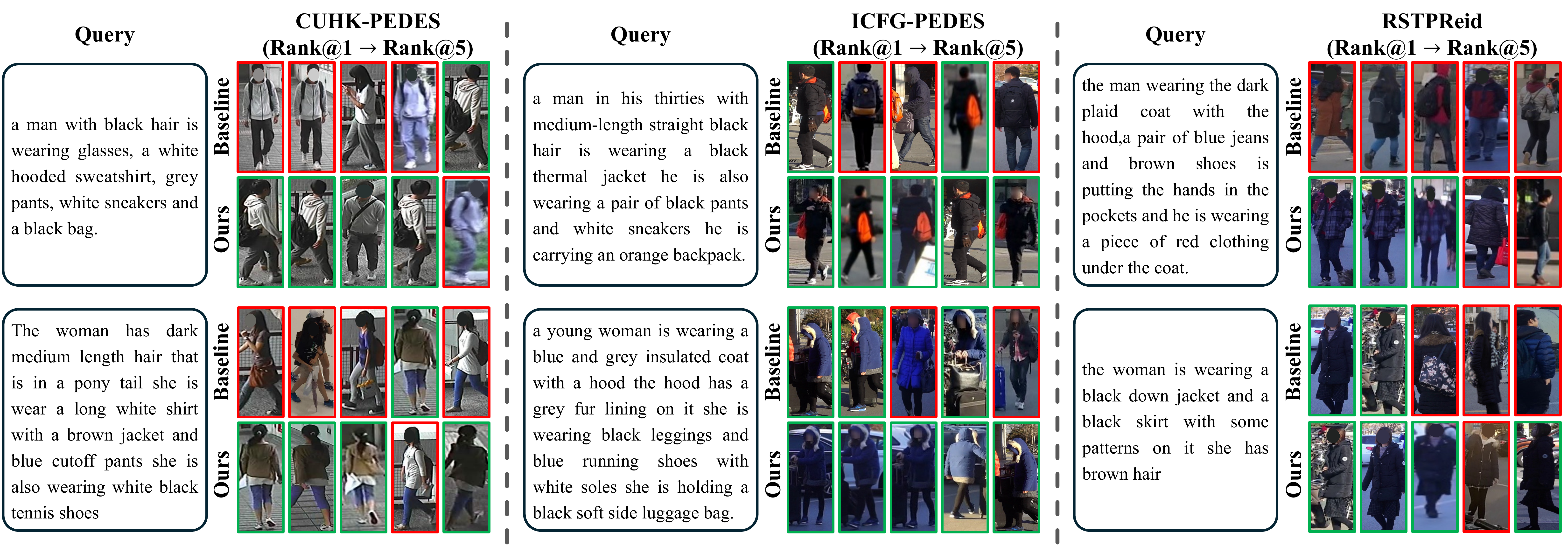}
    \caption{Visualization of retrieval results (R@1 to R@5). Each row represents a query-response group. Green and red bounding boxes indicate correct and incorrect matches, respectively. BLIP~\cite{li2022blip} is used as the baseline.}
  \label{fig:rank}
\end{figure*}

\begin{figure}[htbp]
    \centering
    \includegraphics[width=0.99\linewidth, height=0.62\linewidth]{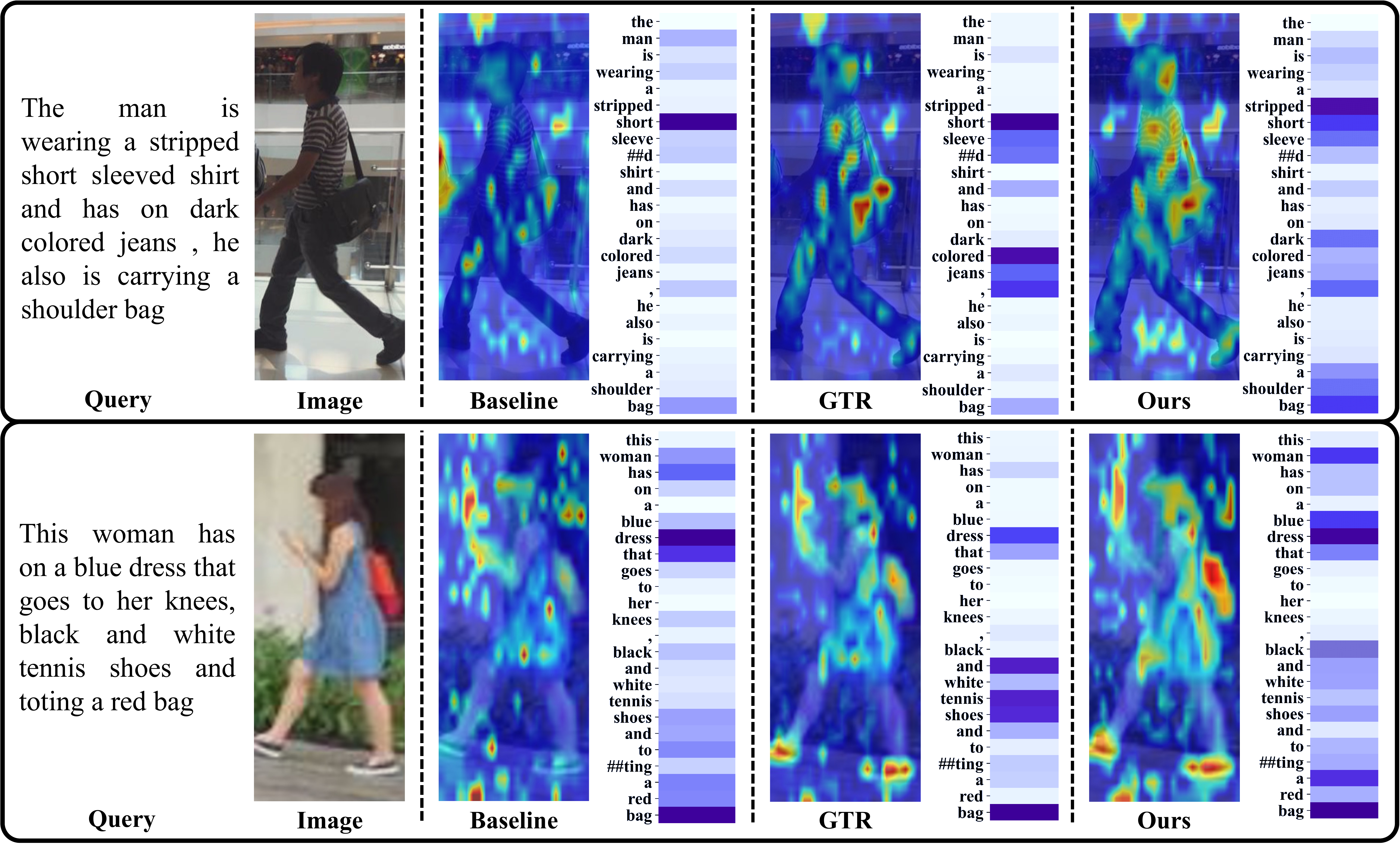}
    \caption{Visualization of activation maps. Warmer colors indicate stronger activation during model computation. BLIP~\cite{li2022blip} is used as the baseline.}
    \label{fig:attnmap}
\end{figure}

Figure~\ref{fig:rank} presents diverse examples illustrating the top-$5$ retrieval results of both the BLIP baseline and our GTR+, demonstrating effectiveness and superiority of GTR+.
Figure~\ref{fig:attnmap} visualizes the activation maps for positive image-text pairs using BLIP~\cite{li2022blip}, GTR~\cite{bai2023text} and our GTR+.
Despite the unsupervised setting, GTR+ places greater emphasis on the described object in the image and the key attributes in the text during computation, compared to other methods.

\section{Conclusion}
In this paper, we address the unsupervised TBPS task, aiming to remove reliance on expensive human-annotated textual descriptions. To this end, we propose GTR+, a two-stage framework that first generates pseudo-descriptions to compensate for the absence of annotations, and subsequently trains the retrieval model in a supervised fashion.
In the generation stage, we introduce a tiered description generation framework that produces fine-grained and stylistically diverse captions through a three-tiered sequential process. 
In the retrieval stage, we propose an adaptive confidence-weighted learning framework that dynamically adjusts each sample's training contribution based on its confidence score, effectively reducing the negative impact of noisy pseudo texts. 
Moreover, we contribute LargeFine-Person, a new large-scale dataset featuring high-quality, fine-grained, and diverse textual annotations, which serves as a more realistic and challenging benchmark for TBPS.
Extensive experiments demonstrate that GTR+ achieves impressive performance on multiple TBPS benchmarks under the unsupervised setting, and also validate the effectiveness of the proposed dataset.


\bibliographystyle{IEEEtran}
\bibliography{ref}

@inproceedings{li2017person,
  title={Person search with natural language description},
  author={Li, Shuang and Xiao, Tong and Li, Hongsheng and Zhou, Bolei and Yue, Dayu and Wang, Xiaogang},
  booktitle={Proceedings of the IEEE conference on computer vision and pattern recognition},
  pages={1970--1979},
  year={2017}
}

@inproceedings{wang2020vitaa,
  title={Vitaa: Visual-textual attributes alignment in person search by natural language},
  author={Wang, Zhe and Fang, Zhiyuan and Wang, Jun and Yang, Yezhou},
  booktitle={Computer Vision--ECCV 2020: 16th European Conference, Glasgow, UK, August 23--28, 2020, Proceedings, Part XII 16},
  pages={402--420},
  year={2020},
  organization={Springer}
}

@inproceedings{zhu2021dssl,
  title={DSSL: Deep Surroundings-person Separation Learning for Text-based Person Retrieval},
  author={Zhu, Aichun and Wang, Zijie and Li, Yifeng and Wan, Xili and Jin, Jing and Wang, Tian and Hu, Fangqiang and Hua, Gang},
  booktitle={Proceedings of the 29th ACM International Conference on Multimedia},
  pages={209--217},
  year={2021}
}

@article{ji2022asymmetric,
  title={Asymmetric Cross-Scale Alignment for Text-Based Person Search},
  author={Ji, Zhong and Hu, Junhua and Liu, Deyin and Wu, Lin Yuanbo and Zhao, Ye},
  journal={IEEE Transactions on Multimedia},
  year={2022},
  publisher={IEEE}
}

@inproceedings{jing2020cross,
  title={Cross-modal cross-domain moment alignment network for person search},
  author={Jing, Ya and Wang, Wei and Wang, Liang and Tan, Tieniu},
  booktitle={Proceedings of the IEEE/CVF Conference on Computer Vision and Pattern Recognition},
  pages={10678--10686},
  year={2020}
}

@inproceedings{li2021unsupervised,
  title={Unsupervised Vision-and-Language Pre-training Without Parallel Images and Captions},
  author={Li, Liunian Harold and You, Haoxuan and Wang, Zhecan and Zareian, Alireza and Chang, Shih-Fu and Chang, Kai-Wei},
  booktitle={Proceedings of the 2021 Conference of the North American Chapter of the Association for Computational Linguistics: Human Language Technologies},
  pages={5339--5350},
  year={2021}
}

@inproceedings{zhou2022unsupervised,
  title={Unsupervised vision-and-language pre-training via retrieval-based multi-granular alignment},
  author={Zhou, Mingyang and Yu, Licheng and Singh, Amanpreet and Wang, Mengjiao and Yu, Zhou and Zhang, Ning},
  booktitle={Proceedings of the IEEE/CVF Conference on Computer Vision and Pattern Recognition},
  pages={16485--16494},
  year={2022}
}

@inproceedings{chen2022end,
  title={End-to-End Unsupervised Vision-and-Language Pre-training with Referring Expression Matching},
  author={Chen, Chi and Li, Peng and Sun, Maosong and Liu, Yang},
  booktitle={Proceedings of the 2022 Conference on Empirical Methods in Natural Language Processing},
  pages={10799--10810},
  year={2022}
}

@inproceedings{feng2019unsupervised,
  title={Unsupervised image captioning},
  author={Feng, Yang and Ma, Lin and Liu, Wei and Luo, Jiebo},
  booktitle={Proceedings of the IEEE/CVF Conference on Computer Vision and Pattern Recognition},
  pages={4125--4134},
  year={2019}
}

@inproceedings{wang2022vlmixer,
  title={Vlmixer: Unpaired vision-language pre-training via cross-modal cutmix},
  author={Wang, Teng and Jiang, Wenhao and Lu, Zhichao and Zheng, Feng and Cheng, Ran and Yin, Chengguo and Luo, Ping},
  booktitle={International Conference on Machine Learning},
  pages={22680--22690},
  year={2022},
  organization={PMLR}
}

@article{dong2021unsupervised,
  title={Unsupervised text-to-image synthesis},
  author={Dong, Yanlong and Zhang, Ying and Ma, Lin and Wang, Zhi and Luo, Jiebo},
  journal={Pattern Recognition},
  volume={110},
  pages={107573},
  year={2021},
  publisher={Elsevier}
}

@inproceedings{guo2021recurrent,
  title={Recurrent relational memory network for unsupervised image captioning},
  author={Guo, Dan and Wang, Yang and Song, Peipei and Wang, Meng},
  booktitle={Proceedings of the Twenty-Ninth International Conference on International Joint Conferences on Artificial Intelligence},
  pages={920--926},
  year={2021}
}

@article{ben2021unpaired,
  title={Unpaired image captioning with semantic-constrained self-learning},
  author={Ben, Huixia and Pan, Yingwei and Li, Yehao and Yao, Ting and Hong, Richang and Wang, Meng and Mei, Tao},
  journal={IEEE Transactions on Multimedia},
  volume={24},
  pages={904--916},
  year={2021},
  publisher={IEEE}
}

@inproceedings{bai2023text,
  title={Text-based person search without parallel image-text data},
  author={Bai, Yang and Wang, Jingyao and Cao, Min and Chen, Chen and Cao, Ziqiang and Nie, Liqiang and Zhang, Min},
  booktitle={Proceedings of the 31st ACM International Conference on Multimedia},
  pages={757--767},
  year={2023}
}

@inproceedings{vinyals2015show,
  title={Show and tell: A neural image caption generator},
  author={Vinyals, Oriol and Toshev, Alexander and Bengio, Samy and Erhan, Dumitru},
  booktitle={Proceedings of the IEEE conference on computer vision and pattern recognition},
  pages={3156--3164},
  year={2015}
}

@article{zheng2020dual,
  title={Dual-path convolutional image-text embeddings with instance loss},
  author={Zheng, Zhedong and Zheng, Liang and Garrett, Michael and Yang, Yi and Xu, Mingliang and Shen, Yi-Dong},
  journal={ACM Transactions on Multimedia Computing, Communications, and Applications (TOMM)},
  volume={16},
  number={2},
  pages={1--23},
  year={2020},
  publisher={ACM New York, NY, USA}
}

@inproceedings{jing2020pose,
  title={Pose-guided multi-granularity attention network for text-based person search},
  author={Jing, Ya and Si, Chenyang and Wang, Junbo and Wang, Wei and Wang, Liang and Tan, Tieniu},
  booktitle={Proceedings of the AAAI Conference on Artificial Intelligence},
  volume={34},
  number={07},
  pages={11189--11196},
  year={2020}
}

@article{niu2020improving,
  title={Improving description-based person re-identification by multi-granularity image-text alignments},
  author={Niu, Kai and Huang, Yan and Ouyang, Wanli and Wang, Liang},
  journal={IEEE Transactions on Image Processing},
  volume={29},
  pages={5542--5556},
  year={2020},
  publisher={IEEE}
}

@article{zhu2023prompt,
  title={Prompt-based learning for unpaired image captioning},
  author={Zhu, Peipei and Wang, Xiao and Zhu, Lin and Sun, Zhenglong and Zheng, Wei-Shi and Wang, Yaowei and Chen, Changwen},
  journal={IEEE Transactions on Multimedia},
  volume={26},
  pages={379--393},
  year={2023},
  publisher={IEEE}
}

@inproceedings{qi2024relational,
  title={Relational distant supervision for image captioning without image-text pairs},
  author={Qi, Yayun and Zhao, Wentian and Wu, Xinxiao},
  booktitle={Proceedings of the AAAI Conference on Artificial Intelligence},
  volume={38},
  number={5},
  pages={4524--4532},
  year={2024}
}

@article{huang2022mack,
  title={MACK: Multimodal aligned conceptual knowledge for unpaired image-text matching},
  author={Huang, Yan and Wang, Yuming and Zeng, Yunan and Wang, Liang},
  journal={Advances in Neural Information Processing Systems},
  volume={35},
  pages={7892--7904},
  year={2022}
}

@inproceedings{zeng2024meacap,
  title={Meacap: Memory-augmented zero-shot image captioning},
  author={Zeng, Zequn and Xie, Yan and Zhang, Hao and Chen, Chiyu and Chen, Bo and Wang, Zhengjue},
  booktitle={Proceedings of the IEEE/CVF conference on computer vision and pattern recognition},
  pages={14100--14110},
  year={2024}
}

@article{sun2024adaptive,
  title={An adaptive correlation filtering method for text-based person search},
  author={Sun, Mengyang and Suo, Wei and Wang, Peng and Niu, Kai and Liu, Le and Lin, Guosheng and Zhang, Yanning and Wu, Qi},
  journal={International Journal of Computer Vision},
  volume={132},
  number={10},
  pages={4440--4455},
  year={2024},
  publisher={Springer}
}

@article{li2023decap,
  title={Decap: Decoding clip latents for zero-shot captioning via text-only training},
  author={Li, Wei and Zhu, Linchao and Wen, Longyin and Yang, Yi},
  journal={arXiv preprint arXiv:2303.03032},
  year={2023}
}

@article{huang2024unpaired,
  title={Unpaired Image-text Matching via Multimodal Aligned Conceptual Knowledge},
  author={Huang, Yan and Wang, Yuming and Zeng, Yunan and Huang, Junshi and Chai, Zhenhua and Wang, Liang},
  journal={IEEE Transactions on Pattern Analysis and Machine Intelligence},
  year={2024},
  publisher={IEEE}
}

@inproceedings{zheng2020hierarchical,
  title={Hierarchical gumbel attention network for text-based person search},
  author={Zheng, Kecheng and Liu, Wu and Liu, Jiawei and Zha, Zheng-Jun and Mei, Tao},
  booktitle={Proceedings of the 28th ACM International Conference on Multimedia},
  pages={3441--3449},
  year={2020}
}

@article{gao2021contextual,
  title={Contextual non-local alignment over full-scale representation for text-based person search},
  author={Gao, Chenyang and Cai, Guanyu and Jiang, Xinyang and Zheng, Feng and Zhang, Jun and Gong, Yifei and Peng, Pai and Guo, Xiaowei and Sun, Xing},
  journal={arXiv preprint arXiv:2101.03036},
  year={2021}
}

@article{liu2025synthesize,
  title={Synthesize then align: Modality alignment augmentation for zero-shot image captioning with synthetic data},
  author={Liu, Zhiyue and Liu, Jinyuan and Ling, Xin and Huang, Qingbao and Wang, Jiahai},
  journal={Knowledge-Based Systems},
  volume={315},
  pages={113274},
  year={2025},
  publisher={Elsevier}
}

@article{gao2022conditional,
  title={Conditional Feature Learning Based Transformer for Text-Based Person Search},
  author={Gao, Chenyang and Cai, Guanyu and Jiang, Xinyang and Zheng, Feng and Zhang, Jun and Gong, Yifei and Lin, Fangzhou and Sun, Xing and Bai, Xiang},
  journal={IEEE Transactions on Image Processing},
  volume={31},
  pages={6097--6108},
  year={2022},
  publisher={IEEE}
}

@article{ding2021semantically,
  title={Semantically self-aligned network for text-to-image part-aware person re-identification},
  author={Ding, Zefeng and Ding, Changxing and Shao, Zhiyin and Tao, Dacheng},
  journal={arXiv preprint arXiv:2107.12666},
  year={2021}
}

@inproceedings{luo2024unleashing,
  title={Unleashing Text-to-Image Diffusion Prior for Zero-Shot Image Captioning},
  author={Luo, Jianjie and Chen, Jingwen and Li, Yehao and Pan, Yingwei and Feng, Jianlin and Chao, Hongyang and Yao, Ting},
  booktitle={European Conference on Computer Vision},
  pages={237--254},
  year={2024},
  organization={Springer}
}

@inproceedings{suo2022simple,
  title={A Simple and Robust Correlation Filtering Method for Text-Based Person Search},
  author={Suo, Wei and Sun, Mengyang and Niu, Kai and Gao, Yiqi and Wang, Peng and Zhang, Yanning and Wu, Qi},
  booktitle={Computer Vision--ECCV 2022: 17th European Conference, Tel Aviv, Israel, October 23--27, 2022, Proceedings, Part XXXV},
  pages={726--742},
  year={2022},
  organization={Springer}
}

@article{han2021text,
  title={Text-based person search with limited data},
  author={Han, Xiao and He, Sen and Zhang, Li and Xiang, Tao},
  journal={arXiv preprint arXiv:2110.10807},
  year={2021}
}

@inproceedings{li2022learning,
  title={Learning semantic-aligned feature representation for text-based person search},
  author={Li, Shiping and Cao, Min and Zhang, Min},
  booktitle={ICASSP 2022-2022 IEEE International Conference on Acoustics, Speech and Signal Processing (ICASSP)},
  pages={2724--2728},
  year={2022},
  organization={IEEE}
}

@article{xu2025sa,
  title={SA-Person: Text-Based Person Retrieval with Scene-aware Re-ranking},
  author={Xu, Yingjia and Wu, Jinlin and Chen, Zhen and Gao, Daming and Yang, Yang and Lei, Zhen and Cao, Min},
  journal={arXiv preprint arXiv:2505.24466},
  year={2025}
}

@inproceedings{wang2022caibc,
  title={CAIBC: Capturing All-round Information Beyond Color for Text-based Person Retrieval},
  author={Wang, Zijie and Zhu, Aichun and Xue, Jingyi and Wan, Xili and Liu, Chao and Wang, Tian and Li, Yifeng},
  booktitle={Proceedings of the 30th ACM International Conference on Multimedia},
  pages={5314--5322},
  year={2022}
}

@article{chen2022tipcb,
  title={TIPCB: A simple but effective part-based convolutional baseline for text-based person search},
  author={Chen, Yuhao and Zhang, Guoqing and Lu, Yujiang and Wang, Zhenxing and Zheng, Yuhui},
  journal={Neurocomputing},
  volume={494},
  pages={171--181},
  year={2022},
  publisher={Elsevier}
}

@inproceedings{liu2024improving,
  title={Improving cross-modal alignment with synthetic pairs for text-only image captioning},
  author={Liu, Zhiyue and Liu, Jinyuan and Ma, Fanrong},
  booktitle={Proceedings of the AAAI Conference on Artificial Intelligence},
  volume={38},
  number={4},
  pages={3864--3872},
  year={2024}
}

@inproceedings{shao2022learning,
  title={Learning Granularity-Unified Representations for Text-to-Image Person Re-identification},
  author={Shao, Zhiyin and Zhang, Xinyu and Fang, Meng and Lin, Zhifeng and Wang, Jian and Ding, Changxing},
  booktitle={Proceedings of the 30th ACM International Conference on Multimedia},
  pages={5566--5574},
  year={2022}
}

@article{zheng2024cpcl,
  title={CPCL: Cross-modal prototypical contrastive learning for weakly supervised text-based person re-identification},
  author={Zheng, Yanwei and Zhao, Xinpeng and Lan, Chuanlin and Zhang, Xiaowei and Huang, Bowen and Yang, Jibin and Yu, Dongxiao},
  journal={arXiv preprint arXiv:2401.10011},
  year={2024}
}

@inproceedings{jiang2023cross,
  title={Cross-modal implicit relation reasoning and aligning for text-to-image person retrieval},
  author={Jiang, Ding and Ye, Mang},
  booktitle={Proceedings of the IEEE/CVF conference on computer vision and pattern recognition},
  pages={2787--2797},
  year={2023}
}

@article{yan2022clip,
  title={CLIP-Driven Fine-grained Text-Image Person Re-identification},
  author={Yan, Shuanglin and Dong, Neng and Zhang, Liyan and Tang, Jinhui},
  journal={arXiv preprint arXiv:2210.10276},
  year={2022}
}

@article{wei2023calibrating,
  title={Calibrating Cross-modal Feature for Text-Based Person Searching},
  author={Wei, Donglai and Zhang, Sipeng and Yang, Tong and Liu, Jing},
  journal={arXiv preprint arXiv:2304.02278},
  year={2023}
}

@article{zhao2021incremental,
  title={Incremental generative occlusion adversarial suppression network for person ReID},
  author={Zhao, Cairong and Lv, Xinbi and Dou, Shuguang and Zhang, Shanshan and Wu, Jun and Wang, Liang},
  journal={IEEE Transactions on Image Processing},
  volume={30},
  pages={4212--4224},
  year={2021},
  publisher={IEEE}
}

@inproceedings{cvpr23crossmodal,
  title={Cross-Modal Implicit Relation Reasoning and Aligning for Text-to-Image Person Retrieval},
  author={Jiang, Ding and Ye, Mang},
  booktitle={IEEE International Conference on Computer Vision and Pattern Recognition (CVPR)},
  year={2023},
}

@inproceedings{bai2025chat,
  title={Chat-based Person Retrieval via Dialogue-Refined Cross-Modal Alignment},
  author={Bai, Yang and Ji, Yucheng and Cao, Min and Wang, Jinqiao and Ye, Mang},
  booktitle={Proceedings of the Computer Vision and Pattern Recognition Conference},
  pages={3952--3962},
  year={2025}
}

@inproceedings{qin2024noisy,
  title={Noisy-correspondence learning for text-to-image person re-identification},
  author={Qin, Yang and Chen, Yingke and Peng, Dezhong and Peng, Xi and Zhou, Joey Tianyi and Hu, Peng},
  booktitle={Proceedings of the IEEE/CVF Conference on Computer Vision and Pattern Recognition},
  pages={27197--27206},
  year={2024}
}

@inproceedings{qu2021dynamic,
  title={Dynamic modality interaction modeling for image-text retrieval},
  author={Qu, Leigang and Liu, Meng and Wu, Jianlong and Gao, Zan and Nie, Liqiang},
  booktitle={Proceedings of the 44th International ACM SIGIR Conference on Research and Development in Information Retrieval},
  pages={1104--1113},
  year={2021}
}

@article{cao2022image,
  title={Image-text retrieval: A survey on recent research and development},
  author={Cao, Min and Li, Shiping and Li, Juntao and Nie, Liqiang and Zhang, Min},
  journal={arXiv preprint arXiv:2203.14713},
  year={2022}
}

@article{ye2021deep,
  title={Deep learning for person re-identification: A survey and outlook},
  author={Ye, Mang and Shen, Jianbing and Lin, Gaojie and Xiang, Tao and Shao, Ling and Hoi, Steven CH},
  journal={IEEE transactions on pattern analysis and machine intelligence},
  volume={44},
  number={6},
  pages={2872--2893},
  year={2021},
  publisher={IEEE}
}

@inproceedings{qiu2024mining,
  title={Mining fine-grained image-text alignment for zero-shot captioning via text-only training},
  author={Qiu, Longtian and Ning, Shan and He, Xuming},
  booktitle={Proceedings of the AAAI Conference on Artificial Intelligence},
  volume={38},
  number={5},
  pages={4605--4613},
  year={2024}
}

@article{farooq2020convolutional,
  title={A convolutional baseline for person re-identification using vision and language descriptions},
  author={Farooq, Ammarah and Awais, Muhammad and Yan, Fei and Kittler, Josef and Akbari, Ali and Khalid, Syed Safwan},
  journal={arXiv preprint arXiv:2003.00808},
  year={2020}
}

@article{loper2002nltk,
  title={Nltk: The natural language toolkit},
  author={Loper, Edward and Bird, Steven},
  journal={arXiv preprint cs/0205028},
  year={2002}
}

@inproceedings{wu2021lapscore,
  title={LapsCore: language-guided person search via color reasoning},
  author={Wu, Yushuang and Yan, Zizheng and Han, Xiaoguang and Li, Guanbin and Zou, Changqing and Cui, Shuguang},
  booktitle={Proceedings of the IEEE/CVF International Conference on Computer Vision},
  pages={1624--1633},
  year={2021}
}

@article{yang2020auto,
  title={Auto-encoding and distilling scene graphs for image captioning},
  author={Yang, Xu and Zhang, Hanwang and Cai, Jianfei},
  journal={IEEE transactions on pattern analysis and machine intelligence},
  volume={44},
  number={5},
  pages={2313--2327},
  year={2020},
  publisher={IEEE}
}

@article{yang2021deconfounded,
  title={Deconfounded image captioning: A causal retrospect},
  author={Yang, Xu and Zhang, Hanwang and Cai, Jianfei},
  journal={IEEE Transactions on Pattern Analysis and Machine Intelligence},
  volume={45},
  number={11},
  pages={12996--13010},
  year={2021},
  publisher={IEEE}
}

@article{yao2010i2t,
  title={I2t: Image parsing to text description},
  author={Yao, Benjamin Z and Yang, Xiong and Lin, Liang and Lee, Mun Wai and Zhu, Song-Chun},
  journal={Proceedings of the IEEE},
  volume={98},
  number={8},
  pages={1485--1508},
  year={2010},
  publisher={IEEE}
}

@inproceedings{fei2023transferable,
  title={Transferable decoding with visual entities for zero-shot image captioning},
  author={Fei, Junjie and Wang, Teng and Zhang, Jinrui and He, Zhenyu and Wang, Chengjie and Zheng, Feng},
  booktitle={Proceedings of the IEEE/CVF international conference on computer vision},
  pages={3136--3146},
  year={2023}
}

@inproceedings{radford2021learning,
  title={Learning transferable visual models from natural language supervision},
  author={Radford, Alec and Kim, Jong Wook and Hallacy, Chris and Ramesh, Aditya and Goh, Gabriel and Agarwal, Sandhini and Sastry, Girish and Askell, Amanda and Mishkin, Pamela and Clark, Jack and others},
  booktitle={International conference on machine learning},
  pages={8748--8763},
  year={2021},
  organization={PMLR}
}

@inproceedings{gu2019unpaired,
  title={Unpaired image captioning via scene graph alignments},
  author={Gu, Jiuxiang and Joty, Shafiq and Cai, Jianfei and Zhao, Handong and Yang, Xu and Wang, Gang},
  booktitle={Proceedings of the IEEE/CVF International Conference on Computer Vision},
  pages={10323--10332},
  year={2019}
}

@article{yang2023exploring,
  title={Exploring diverse in-context configurations for image captioning},
  author={Yang, Xu and Wu, Yongliang and Yang, Mingzhuo and Chen, Haokun and Geng, Xin},
  journal={Advances in Neural Information Processing Systems},
  volume={36},
  pages={40924--40943},
  year={2023}
}

@inproceedings{li2022blip,
  title={Blip: Bootstrapping language-image pre-training for unified vision-language understanding and generation},
  author={Li, Junnan and Li, Dongxu and Xiong, Caiming and Hoi, Steven},
  booktitle={International Conference on Machine Learning},
  pages={12888--12900},
  year={2022},
  organization={PMLR}
}

@inproceedings{tan2024harnessing,
  title={Harnessing the power of mllms for transferable text-to-image person reid},
  author={Tan, Wentan and Ding, Changxing and Jiang, Jiayu and Wang, Fei and Zhan, Yibing and Tao, Dapeng},
  booktitle={Proceedings of the IEEE/CVF Conference on Computer Vision and Pattern Recognition},
  pages={17127--17137},
  year={2024}
}

@inproceedings{yang2023towards,
  title={Towards unified text-based person retrieval: A large-scale multi-attribute and language search benchmark},
  author={Yang, Shuyu and Zhou, Yinan and Zheng, Zhedong and Wang, Yaxiong and Zhu, Li and Wu, Yujiao},
  booktitle={Proceedings of the 31st ACM International Conference on Multimedia},
  pages={4492--4501},
  year={2023}
}

@inproceedings{shao2023unified,
  title={Unified pre-training with pseudo texts for text-to-image person re-identification},
  author={Shao, Zhiyin and Zhang, Xinyu and Ding, Changxing and Wang, Jian and Wang, Jingdong},
  booktitle={Proceedings of the IEEE/CVF International Conference on Computer Vision},
  pages={11174--11184},
  year={2023}
}

@article{cornia2022explaining,
  title={Explaining transformer-based image captioning models: An empirical analysis},
  author={Cornia, Marcella and Baraldi, Lorenzo and Cucchiara, Rita},
  journal={AI Communications},
  volume={35},
  number={2},
  pages={111--129},
  year={2022},
  publisher={SAGE Publications Sage UK: London, England}
}

@article{touvron2023llama,
  title={Llama: Open and efficient foundation language models},
  author={Touvron, Hugo and Lavril, Thibaut and Izacard, Gautier and Martinet, Xavier and Lachaux, Marie-Anne and Lacroix, Timoth{\'e}e and Rozi{\`e}re, Baptiste and Goyal, Naman and Hambro, Eric and Azhar, Faisal and others},
  journal={arXiv preprint arXiv:2302.13971},
  year={2023}
}

@inproceedings{yan2024prototypical,
  title={Prototypical prompting for text-to-image person re-identification},
  author={Yan, Shuanglin and Liu, Jun and Dong, Neng and Zhang, Liyan and Tang, Jinhui},
  booktitle={Proceedings of the 32nd ACM International Conference on Multimedia},
  pages={2331--2340},
  year={2024}
}

@article{qi2025granularity,
  title={Granularity-Aware Hyperbolic Representation for Text-based Person Search},
  author={Qi, Chenghuan and Yang, Xi and Wang, Nannan and Gao, Xinbo},
  journal={IEEE Transactions on Information Forensics and Security},
  year={2025},
  publisher={IEEE}
}

@inproceedings{wu2024laip,
  title={LAIP: learning local alignment from image-phrase modeling for text-based person search},
  author={Wu, Yu and Wang, Haiguang and Wu, Mengxia and Cao, Min and Zhang, Min},
  booktitle={2024 IEEE International Conference on Multimedia and Expo (ICME)},
  pages={1--10},
  year={2024},
  organization={IEEE}
}

@inproceedings{qin2025human,
  title={Human-centered Interactive Learning via MLLMs for Text-to-Image Person Re-identification},
  author={Qin, Yang and Chen, Chao and Fu, Zhihang and Peng, Dezhong and Peng, Xi and Hu, Peng},
  booktitle={Proceedings of the Computer Vision and Pattern Recognition Conference},
  pages={14390--14399},
  year={2025}
}

@article{lu2025prompt,
  title={Prompt-guided Transformer and MLLM Interactive Learning for Text-Based Pedestrian Search},
  author={Lu, Zefeng and Lin, Ronghao and Tan, Yap-Peng and Hu, Haifeng},
  journal={IEEE Transactions on Information Forensics and Security},
  year={2025},
  publisher={IEEE}
}

@article{zhang2025dual,
  title={Dual-Granularity Cross-Modal Identity Association for Weakly-Supervised Text-to-Person Image Matching},
  author={Zhang, Yafei and Shang, Yongle and Li, Huafeng},
  journal={arXiv preprint arXiv:2507.06744},
  year={2025}
}

@inproceedings{park2024plot,
  title={Plot: Text-based person search with part slot attention for corresponding part discovery},
  author={Park, Jicheol and Kim, Dongwon and Jeong, Boseung and Kwak, Suha},
  booktitle={European Conference on Computer Vision},
  pages={474--490},
  year={2024},
  organization={Springer}
}

@article{yang2025minimizing,
  title={Minimizing the Pretraining Gap: Domain-aligned Text-Based Person Retrieval},
  author={Yang, Shuyu and Wang, Yaxiong and Li, Yongrui and Zhu, Li and Zheng, Zhedong},
  journal={arXiv preprint arXiv:2507.10195},
  year={2025}
}

@article{oquab2023dinov2,
  title={Dinov2: Learning robust visual features without supervision},
  author={Oquab, Maxime and Darcet, Timoth{\'e}e and Moutakanni, Th{\'e}o and Vo, Huy and Szafraniec, Marc and Khalidov, Vasil and Fernandez, Pierre and Haziza, Daniel and Massa, Francisco and El-Nouby, Alaaeldin and others},
  journal={arXiv preprint arXiv:2304.07193},
  year={2023}
}

@inproceedings{cao2024empirical,
  title={An empirical study of clip for text-based person search},
  author={Cao, Min and Bai, Yang and Zeng, Ziyin and Ye, Mang and Zhang, Min},
  booktitle={Proceedings of the AAAI Conference on Artificial Intelligence},
  volume={38},
  number={1},
  pages={465--473},
  year={2024}
}

@article{bai2023rasa,
  title={Rasa: Relation and sensitivity aware representation learning for text-based person search},
  author={Bai, Yang and Cao, Min and Gao, Daming and Cao, Ziqiang and Chen, Chen and Fan, Zhenfeng and Nie, Liqiang and Zhang, Min},
  journal={arXiv preprint arXiv:2305.13653},
  year={2023}
}

@inproceedings{barraco2023little,
  title={With a little help from your own past: prototypical memory networks for image captioning},
  author={Barraco, Manuele and Sarto, Sara and Cornia, Marcella and Baraldi, Lorenzo and Cucchiara, Rita},
  booktitle={Proceedings of the IEEE/CVF International Conference on Computer Vision},
  pages={3021--3031},
  year={2023}
}

@article{al2025ensemble,
  title={An ensemble model with attention based mechanism for image captioning},
  author={Al Badarneh, Israa and Hammo, Bassam H and Al-Kadi, Omar},
  journal={Computers and Electrical Engineering},
  volume={123},
  pages={110077},
  year={2025},
  publisher={Elsevier}
}

@inproceedings{li2024adaptive,
  title={Adaptive uncertainty-based learning for text-based person retrieval},
  author={Li, Shenshen and He, Chen and Xu, Xing and Shen, Fumin and Yang, Yang and Shen, Heng Tao},
  booktitle={Proceedings of the AAAI Conference on Artificial Intelligence},
  volume={38},
  number={4},
  pages={3172--3180},
  year={2024}
}

@article{gao2024semi,
  title={Semi-supervised Text-based Person Search},
  author={Gao, Daming and Bai, Yang and Cao, Min and Dou, Hao and Ye, Mang and Zhang, Min},
  journal={arXiv preprint arXiv:2404.18106},
  year={2024}
}

@inproceedings{zhao2021weakly,
  title={Weakly supervised text-based person re-identification},
  author={Zhao, Shizhen and Gao, Changxin and Shao, Yuanjie and Zheng, Wei-Shi and Sang, Nong},
  booktitle={Proceedings of the IEEE/CVF international conference on computer vision},
  pages={11395--11404},
  year={2021}
}

@article{hessel2021clipscore,
  title={Clipscore: A reference-free evaluation metric for image captioning},
  author={Hessel, Jack and Holtzman, Ari and Forbes, Maxwell and Bras, Ronan Le and Choi, Yejin},
  journal={arXiv preprint arXiv:2104.08718},
  year={2021}
}

@inproceedings{chen2024sharegpt4v,
  title={Sharegpt4v: Improving large multi-modal models with better captions},
  author={Chen, Lin and Li, Jinsong and Dong, Xiaoyi and Zhang, Pan and He, Conghui and Wang, Jiaqi and Zhao, Feng and Lin, Dahua},
  booktitle={European Conference on Computer Vision},
  pages={370--387},
  year={2024},
  organization={Springer}
}

@inproceedings{li2024cross,
  title={Cross-modal generation and alignment via attribute-guided prompt for unsupervised text-based person retrieval},
  author={Li, Zongyi and Li, Jianbo and Shi, Yuxuan and Ling, Hefei and Chen, Jiazhong and Wang, Runsheng and Huang, Shijuan},
  booktitle={Proceedings of the International Joint Conference on Artificial Intelligence. International Joint Conferences on Artificial Intelligence Organization},
  pages={1047--1055},
  year={2024}
}

@inproceedings{li2025exploring,
  title={Exploring the Potential of Large Vision-Language Models for Unsupervised Text-Based Person Retrieval},
  author={Li, Zongyi and Jianbo, Li and Shi, Yuxuan and Chen, Jiazhong and Huang, Shijuan and Tu, Linnan and Shen, Fei and Ling, Hefei},
  booktitle={Proceedings of the AAAI Conference on Artificial Intelligence},
  volume={39},
  number={5},
  pages={5119--5127},
  year={2025}
}

@inproceedings{gong2024enhancing,
  title={Enhancing cross-modal completion and alignment for unsupervised incomplete text-to-image person retrieval},
  author={Gong, Tiantian and Wang, Junsheng and Zhang, Liyan},
  booktitle={Proceedings of the Thirty-Third International Joint Conference on Artificial Intelligence},
  pages={794--802},
  year={2024}
}

@article{fu2025similarity,
  title={Similarity Regulation and Calibration Alignment for Weakly Supervised Text-Based Person Re-Identification},
  author={Fu, Ao and Zhao, Jiaqi and Zhou, Yong and Du, Wenliang and Yao, Rui and El Saddik, Abdulmotaleb},
  journal={ACM Transactions on Multimedia Computing, Communications and Applications},
  volume={21},
  number={3},
  pages={1--19},
  year={2025},
  publisher={ACM New York, NY}
}

@article{chen2025blip3,
  title={Blip3-o: A family of fully open unified multimodal models-architecture, training and dataset},
  author={Chen, Jiuhai and Xu, Zhiyang and Pan, Xichen and Hu, Yushi and Qin, Can and Goldstein, Tom and Huang, Lifu and Zhou, Tianyi and Xie, Saining and Savarese, Silvio and others},
  journal={arXiv preprint arXiv:2505.09568},
  year={2025}
}

@article{wang2024qwen2,
  title={Qwen2-vl: Enhancing vision-language model's perception of the world at any resolution},
  author={Wang, Peng and Bai, Shuai and Tan, Sinan and Wang, Shijie and Fan, Zhihao and Bai, Jinze and Chen, Keqin and Liu, Xuejing and Wang, Jialin and Ge, Wenbin and others},
  journal={arXiv preprint arXiv:2409.12191},
  year={2024}
}

@article{chen2024expanding,
  title={Expanding performance boundaries of open-source multimodal models with model, data, and test-time scaling},
  author={Chen, Zhe and Wang, Weiyun and Cao, Yue and Liu, Yangzhou and Gao, Zhangwei and Cui, Erfei and Zhu, Jinguo and Ye, Shenglong and Tian, Hao and Liu, Zhaoyang and others},
  journal={arXiv preprint arXiv:2412.05271},
  year={2024}
}

@inproceedings{jiang2025modeling,
  title={Modeling Thousands of Human Annotators for Generalizable Text-to-Image Person Re-identification},
  author={Jiang, Jiayu and Ding, Changxing and Tan, Wentao and Wang, Junhong and Tao, Jin and Xu, Xiangmin},
  booktitle={Proceedings of the Computer Vision and Pattern Recognition Conference},
  pages={9220--9230},
  year={2025}
}

@article{zuo2024plip,
  title={Plip: Language-image pre-training for person representation learning},
  author={Zuo, Jialong and Hong, Jiahao and Zhang, Feng and Yu, Changqian and Zhou, Hanyu and Gao, Changxin and Sang, Nong and Wang, Jingdong},
  journal={Advances in Neural Information Processing Systems},
  volume={37},
  pages={45666--45702},
  year={2024}
}

@article{huang2021learning,
  title={Learning with noisy correspondence for cross-modal matching},
  author={Huang, Zhenyu and Niu, Guocheng and Liu, Xiao and Ding, Wenbiao and Xiao, Xinyan and Wu, Hua and Peng, Xi},
  journal={Advances in Neural Information Processing Systems},
  volume={34},
  pages={29406--29419},
  year={2021}
}

@article{dang2025disentangled,
  title={Disentangled Noisy Correspondence Learning},
  author={Dang, Zhuohang and Luo, Minnan and Wang, Jihong and Jia, Chengyou and Han, Haochen and Wan, Herun and Dai, Guang and Chang, Xiaojun and Wang, Jingdong},
  journal={IEEE Transactions on Image Processing},
  year={2025},
  publisher={IEEE}
}

@inproceedings{han2023noisy,
  title={Noisy correspondence learning with meta similarity correction},
  author={Han, Haochen and Miao, Kaiyao and Zheng, Qinghua and Luo, Minnan},
  booktitle={Proceedings of the IEEE/CVF Conference on Computer Vision and Pattern Recognition},
  pages={7517--7526},
  year={2023}
}

@inproceedings{song2018region,
  title={Region-based quality estimation network for large-scale person re-identification},
  author={Song, Guanglu and Leng, Biao and Liu, Yu and Hetang, Congrui and Cai, Shaofan},
  booktitle={Proceedings of the AAAI conference on artificial intelligence},
  volume={32},
  number={1},
  year={2018}
}

@inproceedings{fu2021unsupervised,
  title={Unsupervised pre-training for person re-identification},
  author={Fu, Dengpan and Chen, Dongdong and Bao, Jianmin and Yang, Hao and Yuan, Lu and Zhang, Lei and Li, Houqiang and Chen, Dong},
  booktitle={Proceedings of the IEEE/CVF conference on computer vision and pattern recognition},
  pages={14750--14759},
  year={2021}
}

@article{shi2024breaking,
  title={Breaking through the noisy correspondence: A robust model for image-text matching},
  author={Shi, Haitao and Liu, Meng and Mu, Xiaoxuan and Song, Xuemeng and Hu, Yupeng and Nie, Liqiang},
  journal={ACM Transactions on Information Systems},
  volume={42},
  number={6},
  pages={1--26},
  year={2024},
  publisher={ACM New York, NY, USA}
}

@inproceedings{bucciarelli2025personalizing,
  title={Personalizing multimodal large language models for image captioning: an Experimental analysis},
  author={Bucciarelli, Davide and Moratelli, Nicholas and Cornia, Marcella and Baraldi, Lorenzo and Cucchiara, Rita},
  booktitle={European Conference on Computer Vision},
  pages={351--368},
  year={2025},
  organization={Springer}
}

@inproceedings{fu2022large,
  title={Large-scale pre-training for person re-identification with noisy labels},
  author={Fu, Dengpan and Chen, Dongdong and Yang, Hao and Bao, Jianmin and Yuan, Lu and Zhang, Lei and Li, Houqiang and Wen, Fang and Chen, Dong},
  booktitle={Proceedings of the IEEE/CVF conference on computer vision and pattern recognition},
  pages={2476--2486},
  year={2022}
}

@article{dong2024benchmarking,
  title={Benchmarking and improving detail image caption},
  author={Dong, Hongyuan and Li, Jiawen and Wu, Bohong and Wang, Jiacong and Zhang, Yuan and Guo, Haoyuan},
  journal={arXiv preprint arXiv:2405.19092},
  year={2024}
}

\vspace{-1cm}
\begin{IEEEbiography}
[{\includegraphics[width=1in, height=1.25in, clip, keepaspectratio]{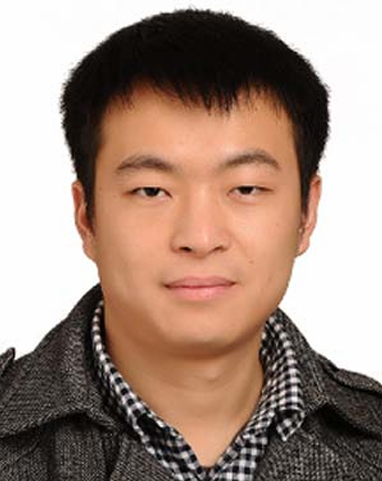}}]
{Mang Ye}
(Senior Member, IEEE) received the PhD degree in computer science from Hong Kong Baptist University, in 2019. He is currently a full professor with the School of Computer Science, Wuhan University, Wuhan, China. He has published more than 100 articles in top-tier venues. He serves as the associate editor for IEEE Transactions on Image Processing, IEEE Transactions on Information Forensics and Security, the Journal of Electronic Imaging, and CAAI Transactions on Intelligence Technology. His research interests focus on computer vision, pattern recognition, and federated learning.
\end{IEEEbiography}
\vspace{-1cm}

\begin{IEEEbiography}
[{\includegraphics[width=1in, height=1.25in, clip, keepaspectratio]{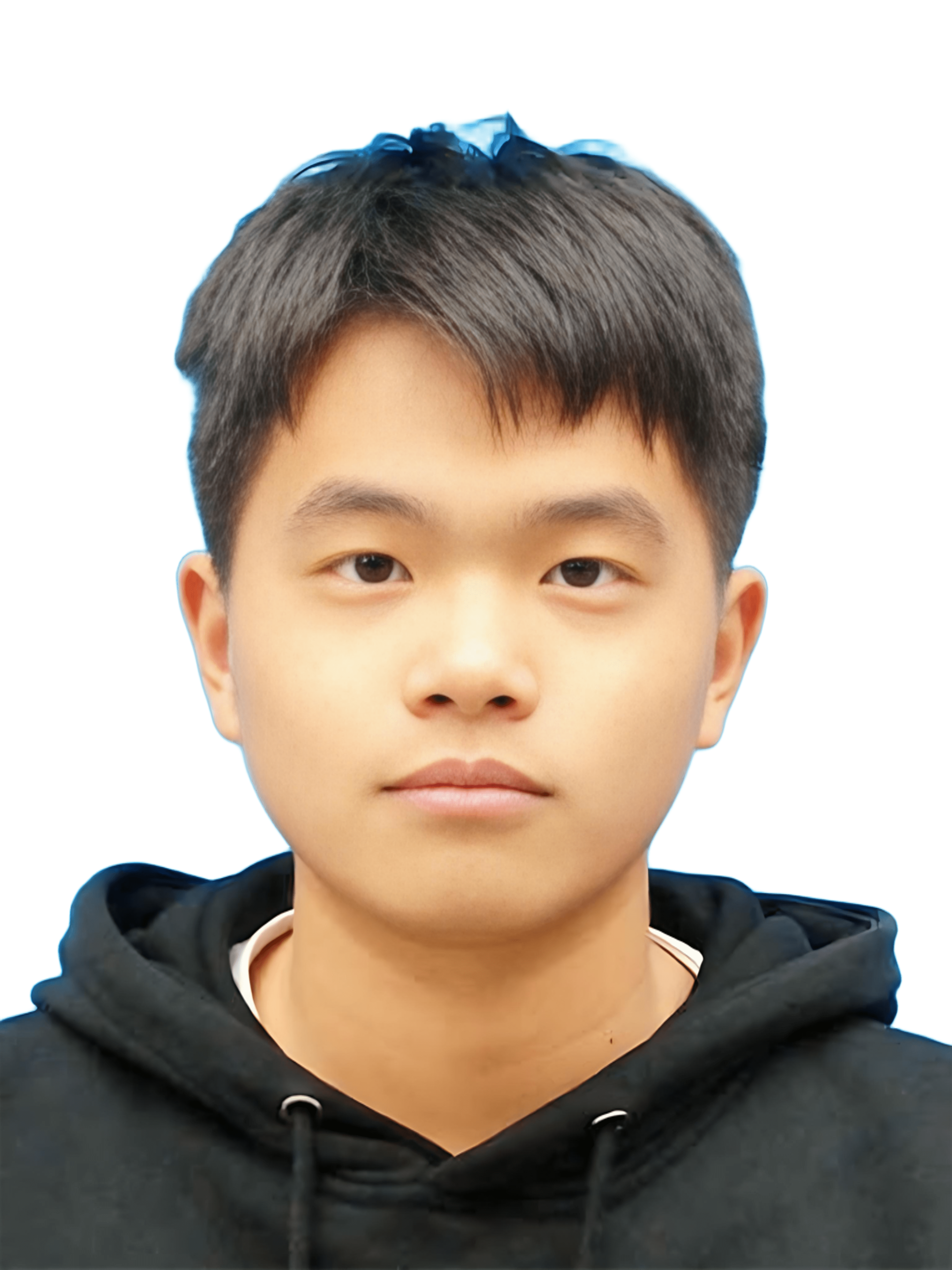}}]
{Yucheng Ji}
received the B.E. degree from the School of Future Science and Engineering, Soochow University, Suzhou, China. He is currently working toward the ME degree with the School of Computer Science and Technology, Soochow University. His research interests include cross-modal retrieval and person re-identification.
\end{IEEEbiography}
\vspace{-1cm}

\begin{IEEEbiography}
[{\includegraphics[width=1in, height=1.25in, clip, keepaspectratio]{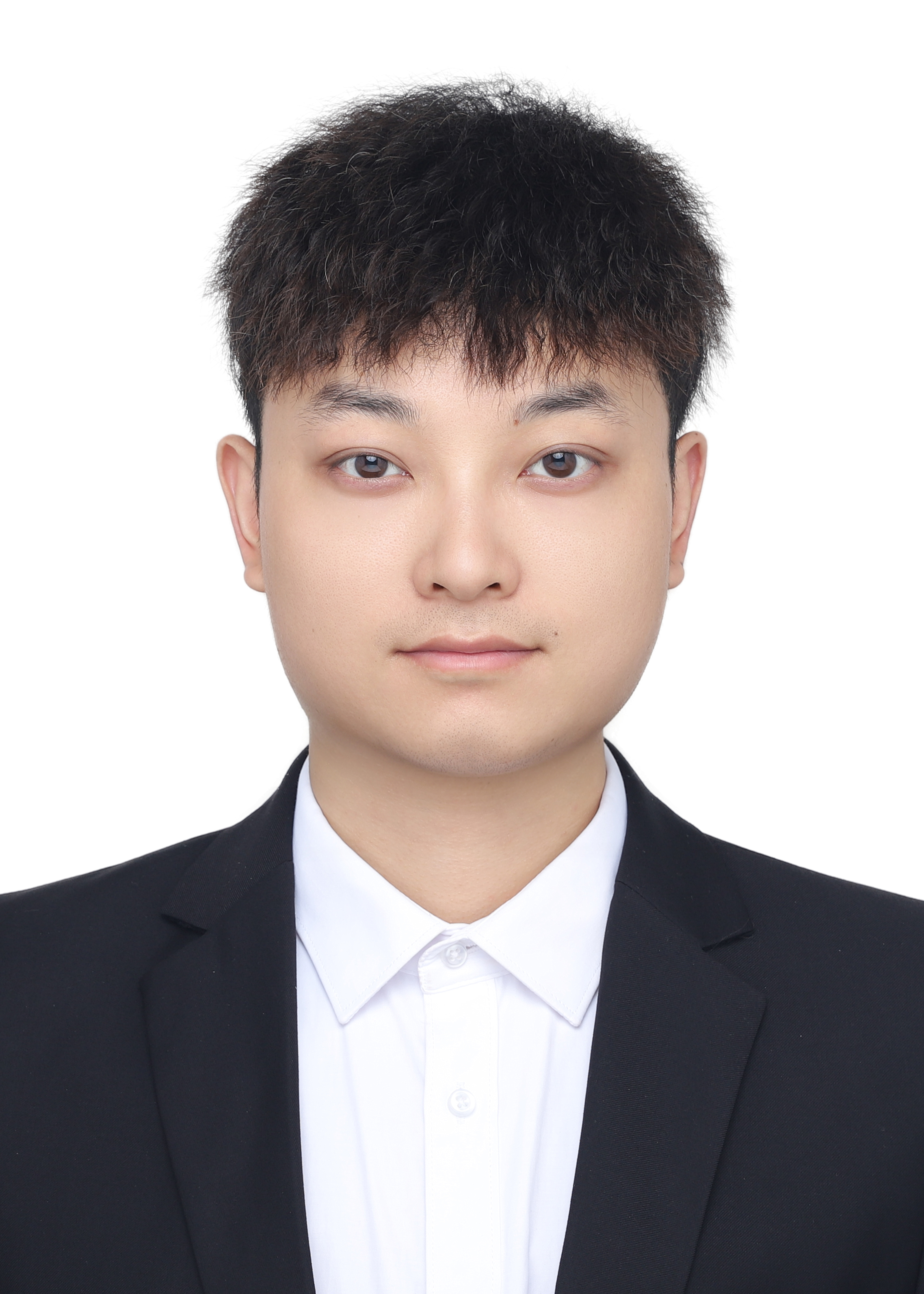}}]
{Yang Bai}
received the M.E. degree from the School of Computer Science and Technology, Soochow University, Suzhou, China. He is currently pursuing the Ph.D. degree with the School of Computer Science, Wuhan University. His research interests include cross-modal retrieval and person re-identification.
\end{IEEEbiography}
\vspace{-1cm}

\begin{IEEEbiography}
[{\includegraphics[width=1in, height=1.25in, clip, keepaspectratio]{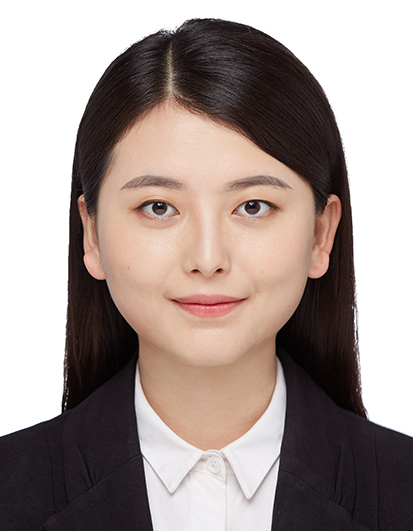}}]
{Min Cao}
received her Ph.D. degree in pattern recognition and intelligent systems from the Institute of Automation, Chinese Academy of Sciences, Beijing, China, in 2020. In March 2020, she became a member of the computer science and technology school at Soochow University, where she is currently an Associate Professor. She was a visiting scholar in computer graphics research, Fraunhofer-Gesellschaft, Darmstadt, Germany, in 2018. Her research interests include cross-modal vision-language learning and person re-identification.
\end{IEEEbiography}
\vspace{-5cm}

\begin{IEEEbiography}
[{\includegraphics[width=1in, height=1.25in, clip, keepaspectratio]{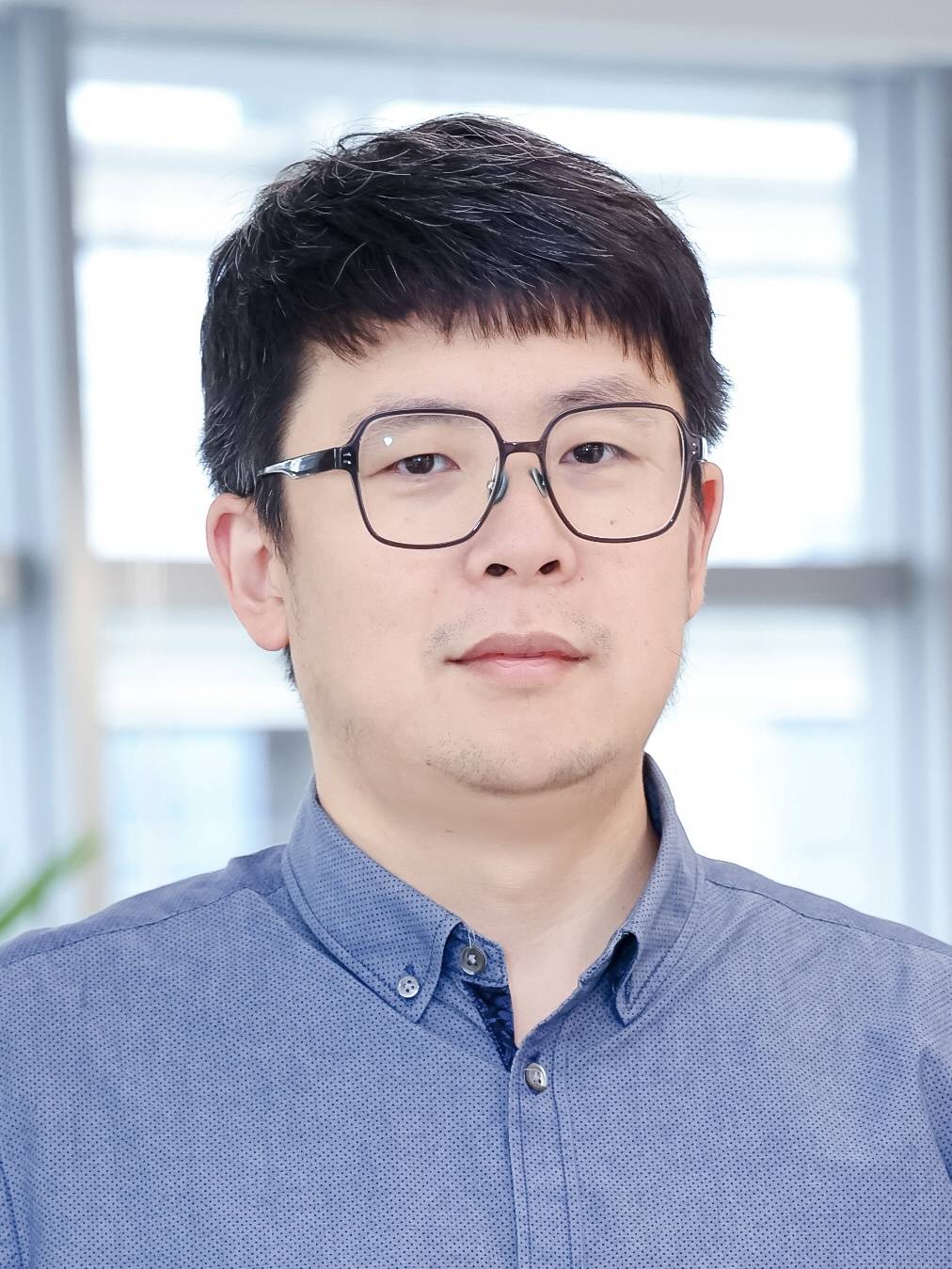}}]
{Siyuan Chai}
received his M.E. degree from the School of Computer Science and Technology, Jilin University, Jilin, China. Since 2023, he has been serving as the General Manager of the Business Technology Center at Zhipu AI. His research interests focus on the applications of large language models, including Retrieval-Augmented Generation (RAG), Agents, Natural Language to SQL (NL2SQL) conversion, and multimodal large language models.
\end{IEEEbiography}
\vspace{-5cm}

\begin{IEEEbiography}
[{\includegraphics[width=1in, height=1.25in, clip, keepaspectratio]{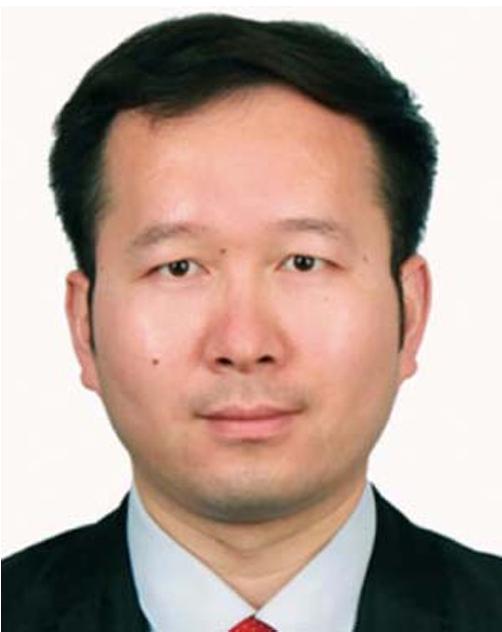}}]
{Bo Du}
(Senior Member, IEEE) received the PhD degree in photogrammetry and remote sensing from the State Key Laboratory of Information Engineering in Surveying, Mapping and Remote Sensing, Wuhan University, Wuhan, China, in 2010. He is a professor with the School of Computer Science, Wuhan University. He has more than 60 research articles published in the IEEE TGRS, TIP, JSTARS, and GRSL. Thirteen of them are ESI hot articles or highly cited articles. His major research interests include pattern recognition, hyperspectral image processing, machine learning, and signal processing. He was a recipient of the Distinguished Paper Award from IJCAI 2018, the Best Paper Award of the IEEE Whispers 2018, the Champion Award of the IEEE Data Fusion Contest 2018, the Best Reviewer Award from the IEEE GRSS for his service to the IEEE Journal of Selected Topics in Earth Observations and Applied Remote Sensing, in 2011, and the ACM rising star awards for his academic progress, in 2015. He was the session chair of the IGARSS 2018/2016 and the 4th IEEE GRSS Workshop on Hyperspectral Image and Signal Processing: Evolution in Remote Sensing.
\end{IEEEbiography}
\vspace{-5cm}

\begin{IEEEbiography}
[{\includegraphics[width=1in, height=1.25in, clip, keepaspectratio]{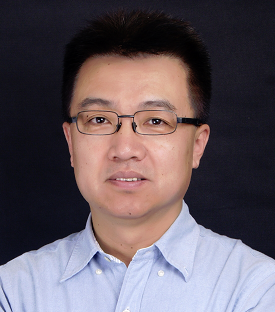}}]
{Min Zhang}
(ACL Fellow, and AAIA Fellow) received the bachelor’s and PhD degrees from the Harbin Institute of Technology, in 1991 and 1997, respectively. He is a distinguished professor with Soochow University (China). His current research interests include machine translation, natural language processing, large model and AI.
\end{IEEEbiography}


\end{document}